\documentclass[letterpaper, 10 pt, conference]{ieeeconf}  % Comment this line out if you need a4paper

\IEEEoverridecommandlockouts                              % This command is only needed if 
\usepackage{amsmath}
\let\labelindent\IEEElabelindent
\usepackage[shortlabels]{enumitem}
\usepackage{algorithm}
\usepackage{algpseudocode}
\usepackage{graphicx}
\usepackage[colorlinks=true, linkcolor=red, citecolor=red, urlcolor=red]{hyperref}

\usepackage{algorithmicx}

\usepackage{subcaption}

\title{\LARGE \bf
An Optimization-Augmented Control Framework for Single and Coordinated Multi-Arm Robotic Manipulation
}

\author{Melih Özcan$^{1}$ and Ozgur S. Oguz$^{2}$% <-this % stops a space
\thanks{*This work was supported by TUBITAK under 2232 program with project number 121C148 (“LiRA”).}% <-this % stops a space
\thanks{$^{1}$Melih Özcan is with Department of Electrical and Electronics Engineering, Middle East Technical University, Ankara, Turkiye.
        {\tt\small melih.ozcan@metu.edu.tr}}%
\thanks{$^{2}$Ozgur S. Oguz is with the Department of Computer Engineering, Bilkent University, Ankara, Turkiye.
        {\tt\small ozgur@cs.bilkent.edu.tr}}%
}

\begin{document}

\maketitle
\thispagestyle{empty}
\pagestyle{empty}

%%%%%%%%%%%%%%%%%%%%%%%%%%%%%%%%%%%%%%%%%%%%%%%%%%%%%%%%%%%%%%%%%%%%%%%%%%%%%%%%
\begin{abstract}

Robotic manipulation demands precise control over both contact forces and motion trajectories. While force control is essential for achieving compliant interaction and high-frequency adaptation, it is limited to operations in close proximity to the manipulated object and often fails to maintain stable orientation during extended motion sequences. Conversely, optimization-based motion planning excels in generating collision-free trajectories over the robot’s configuration space but struggles with dynamic interactions where contact forces play a crucial role. 
To address these limitations, we propose a multi-modal control framework that combines force control and optimization-augmented motion planning to tackle complex robotic manipulation tasks in a sequential manner, enabling seamless switching between control modes based on task requirements. 
Our approach decomposes complex tasks into subtasks, each dynamically assigned to one of three control modes: Pure optimization for global motion planning, pure force control for precise interaction, or hybrid control for tasks requiring simultaneous trajectory tracking and force regulation. 
This framework is particularly advantageous for bimanual and multi-arm manipulation, where synchronous motion and coordination among arms are essential while considering both the manipulated object and environmental constraints. 
We demonstrate the versatility of our method through a range of long-horizon manipulation tasks, including single-arm, bimanual, and multi-arm applications, highlighting its ability to handle both free-space motion and contact-rich manipulation with robustness and precision.
More information is available at \url{https://sites.google.com/view/komo-force/home}
\end{abstract}

%%%%%%%%%%%%%%%%%%%%%%%%%%%%%%%%%%%%%%%%%%%%%%%%%%%%%%%%%%%%%%%%%%%%%%%%%%%%%%%%
\section{INTRODUCTION}

The field of robotics has witnessed remarkable progress in recent years, with robots increasingly being deployed in industrial and household settings. 
%Among these advancements, humanoid robots have garnered significant attention due to their potential to perform complex tasks that require dexterity, adaptability, and human-like coordination. 
For robots to operate effectively in real-world environments, they must not only navigate freely in unstructured spaces but also interact seamlessly with objects and their surroundings. 
A critical aspect of such interactions is the ability to regulate contact forces while maintaining precise motion control—a capability that remains a central challenge in robotic manipulation.

\begin{figure}[thpb]
\label{task1}
\vspace{-4mm}
    \centering
    \begin{subfigure}{0.238\textwidth} % Adjust width
        \centering
        \includegraphics[width=\linewidth]{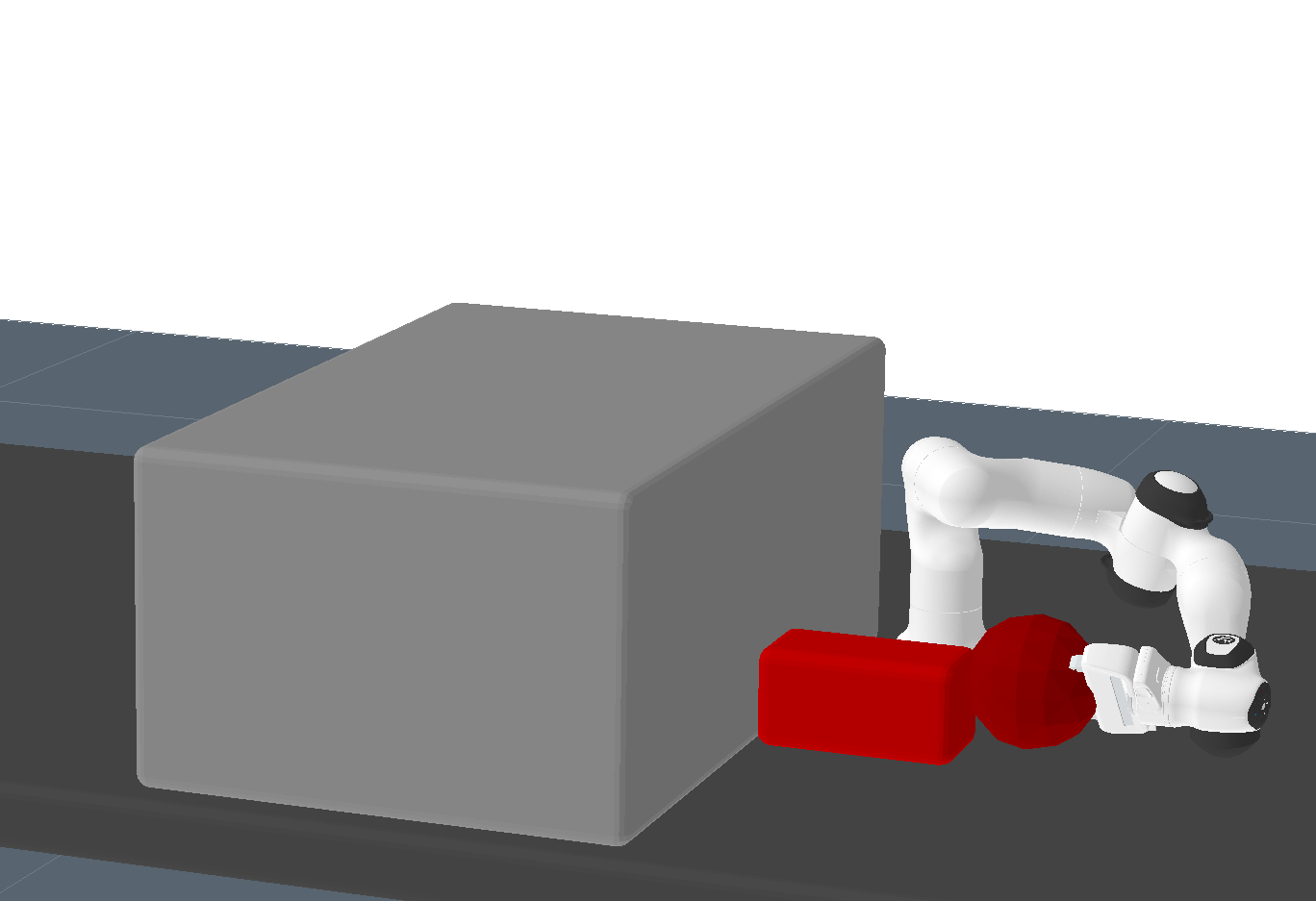}
        \caption{approach the box}
        \label{fig:subfig10}
    \end{subfigure}
    \begin{subfigure}{0.238\textwidth}
        \centering
        \includegraphics[width=\linewidth]{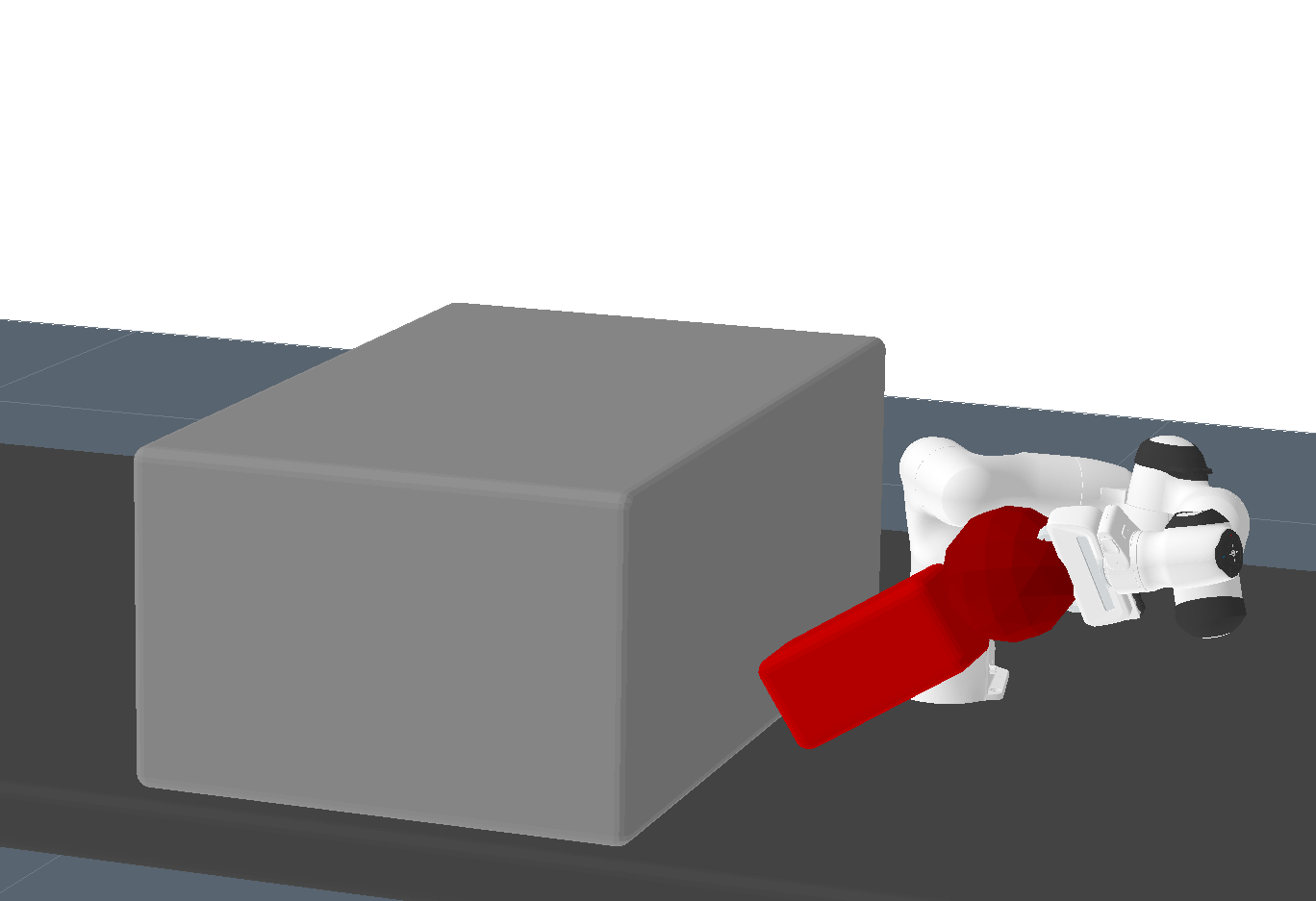}
        \caption{rotate it}
        \label{fig:subfig20}
    \end{subfigure}
    
    % \vspace{0.2cm} % Add vertical space between rows

    \begin{subfigure}{0.238\textwidth}
    \vspace{-6mm}
        \centering
        \includegraphics[width=\linewidth]{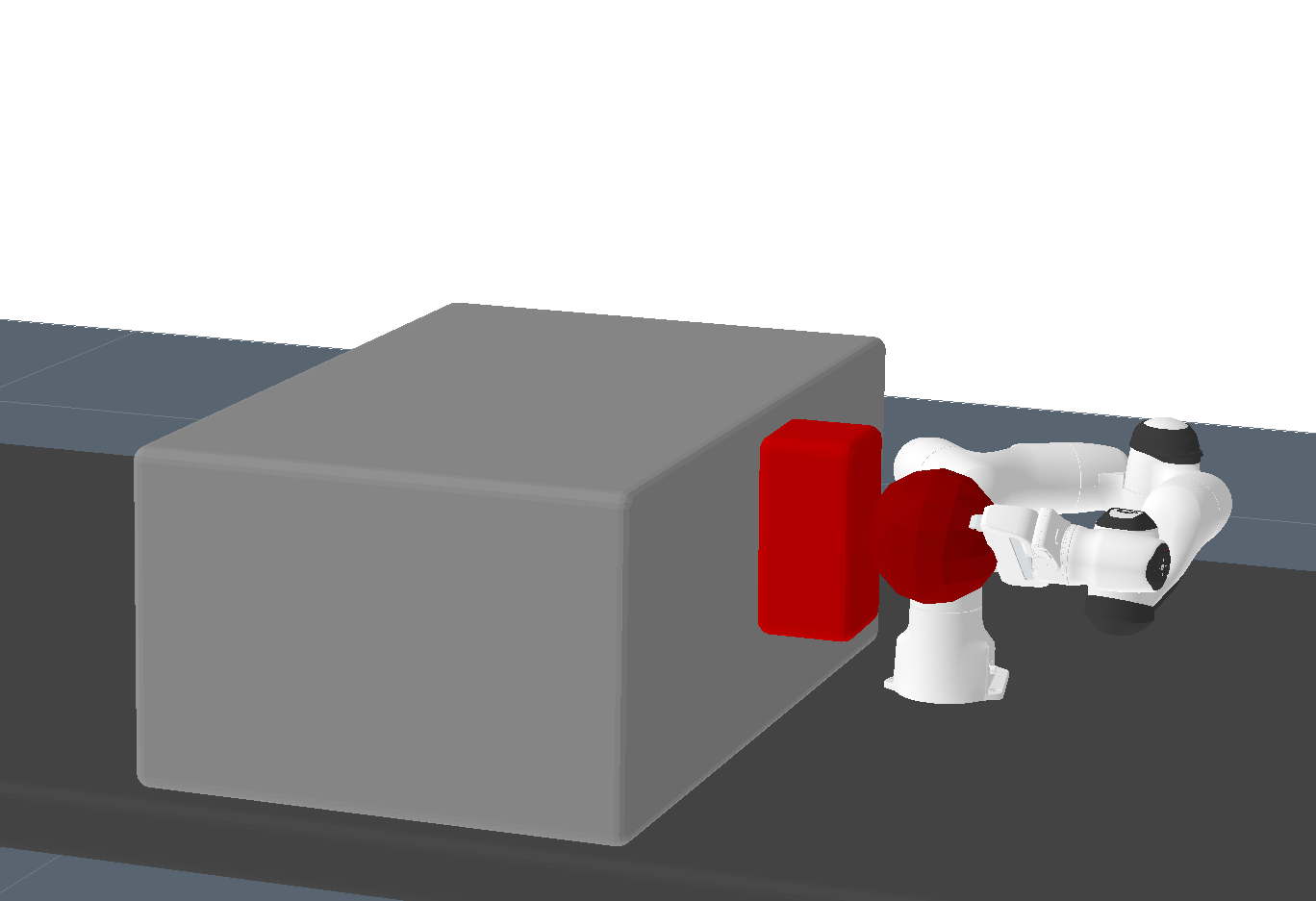}
        \caption{carry the box upward}
        \label{fig:subfig30}
    \end{subfigure}
    \begin{subfigure}{0.238\textwidth}
        \centering
        \includegraphics[width=\linewidth]{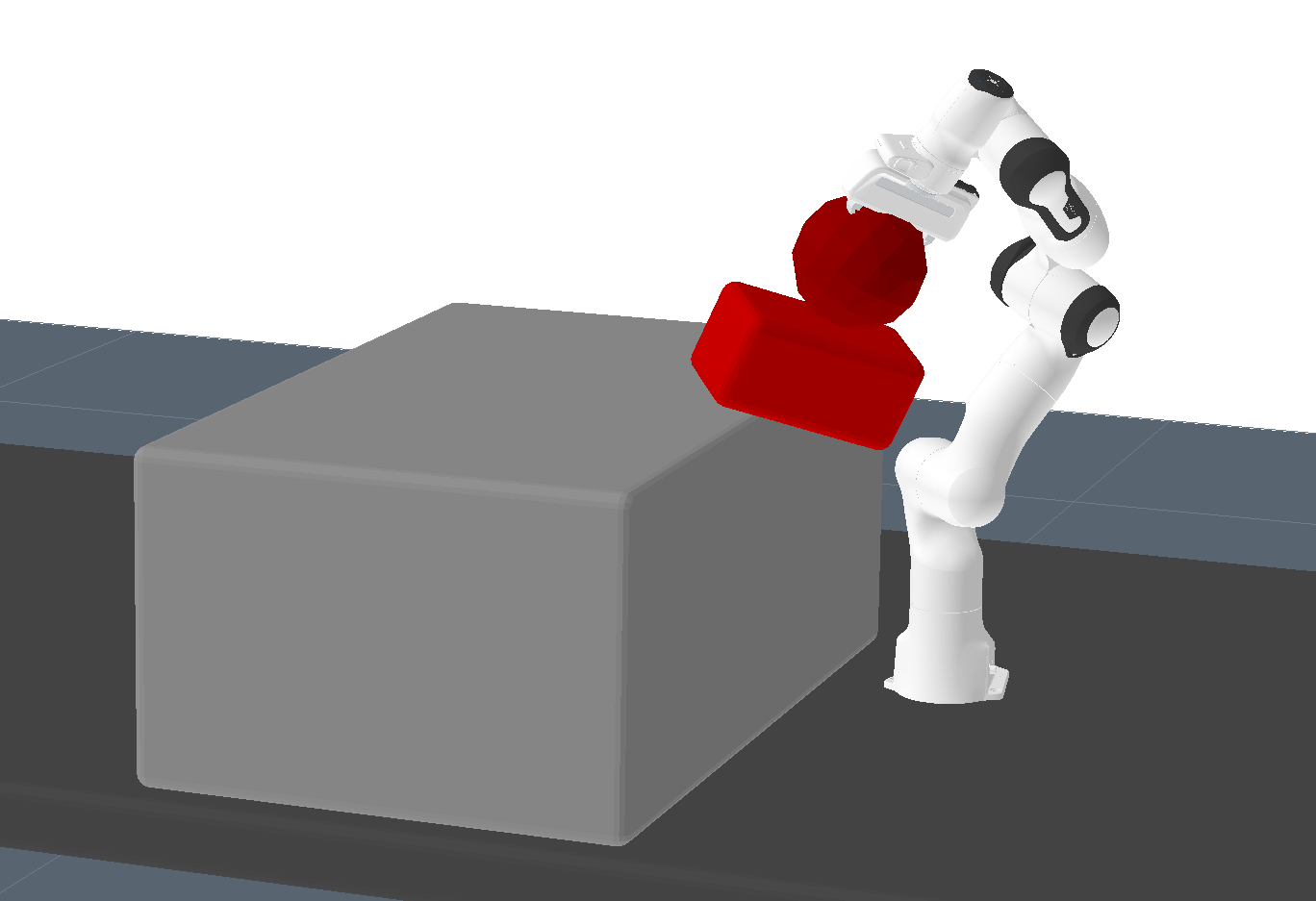}
        \caption{bring it onto the platform}
        \label{fig:subfig40}
    \end{subfigure}

    \caption{Robot approaches the box, rotates it, then carries it onto the platform while keeping its end effector perpendicular to the motion direction.}
    \label{fig:exp1}
\end{figure}

To address this challenge, two dominant paradigms have emerged: optimization-based motion planning and force control. 
Optimization-based methods excel in generating collision-free trajectories for free-space motion or scenarios where contact forces are minimal. 
These approaches optimize paths over the robot's configuration space, ensuring smooth transitions and adherence to constraints such as joint limits and obstacle avoidance. 
However, when contact forces become significant—such as during grasping, pushing, or manipulating delicate objects—optimization-based methods often fall short. 
Their reliance on kinematic planning makes them ill-suited for real-time adaptation to dynamic contact conditions, leading to suboptimal performance in tasks requiring high-frequency responsiveness.

On the other hand, force control provides an effective solution for tasks demanding precise regulation of interaction forces \cite{hybridForce}. By continuously adjusting forces and torques at the end-effector, force control enables compliant and adaptive behavior, making it ideal for operations involving sustained physical contact. Despite its strengths, force control struggles with long-horizon planning and maintaining stable orientations during extended motions. For instance, transporting an object over a long distance using force control alone can result in excessive drift, compromising task success. Furthermore, synchronizing multiple arms—a common requirement in bimanual or multi-arm manipulation—is inherently challenging without higher-level planning.

A deeper examination reveals additional challenges that underscore the limitations of existing paradigms. First, it is generally not possible to simultaneously control position and orientation of a robot arm with force control, as their dynamics are highly coupled. While operational space control (OSC) attempts to achieve this by prioritizing position or orientation, and controlling the other in a way that won't affect the prioritized task, it does not guarantee convergence to both desired position and orientation~\cite{operationalSpaceControl}. 
% Even when operational space control succeeds, it does not take critical constraints such as joint limits into account, which are essential for long-horizon planning. 
Even though OSC can incorporate constraints such as joint limits through null-space projections or optimization-based extensions, standard formulations often do not explicitly account for them. 
This limitation highlights the necessity of optimization-based methods for generating feasible trajectories that consider all relevant constraints.

\begin{figure*}[h]
    \centering
    \includegraphics[width=0.19\textwidth]{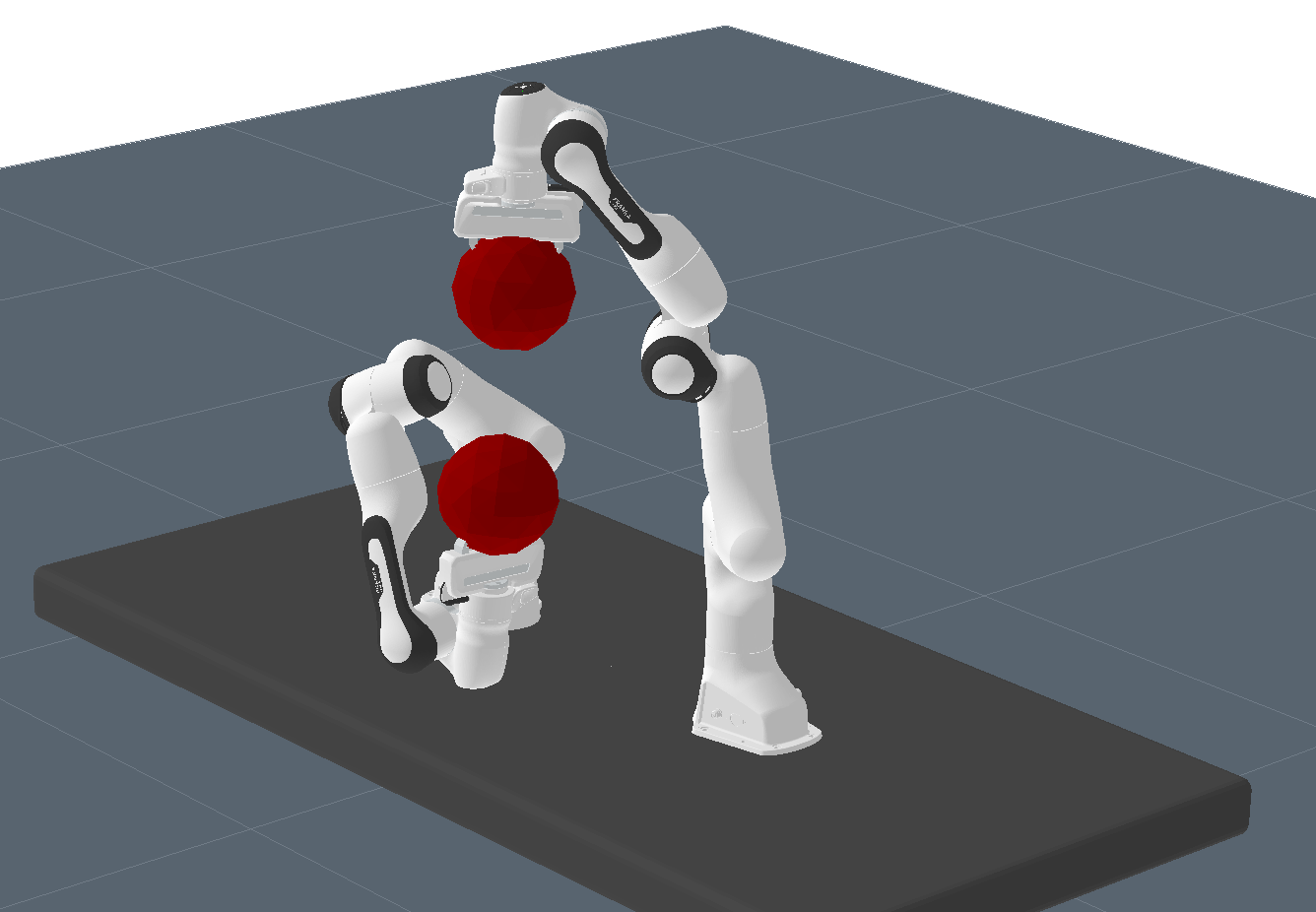}
    \includegraphics[width=0.19\textwidth]{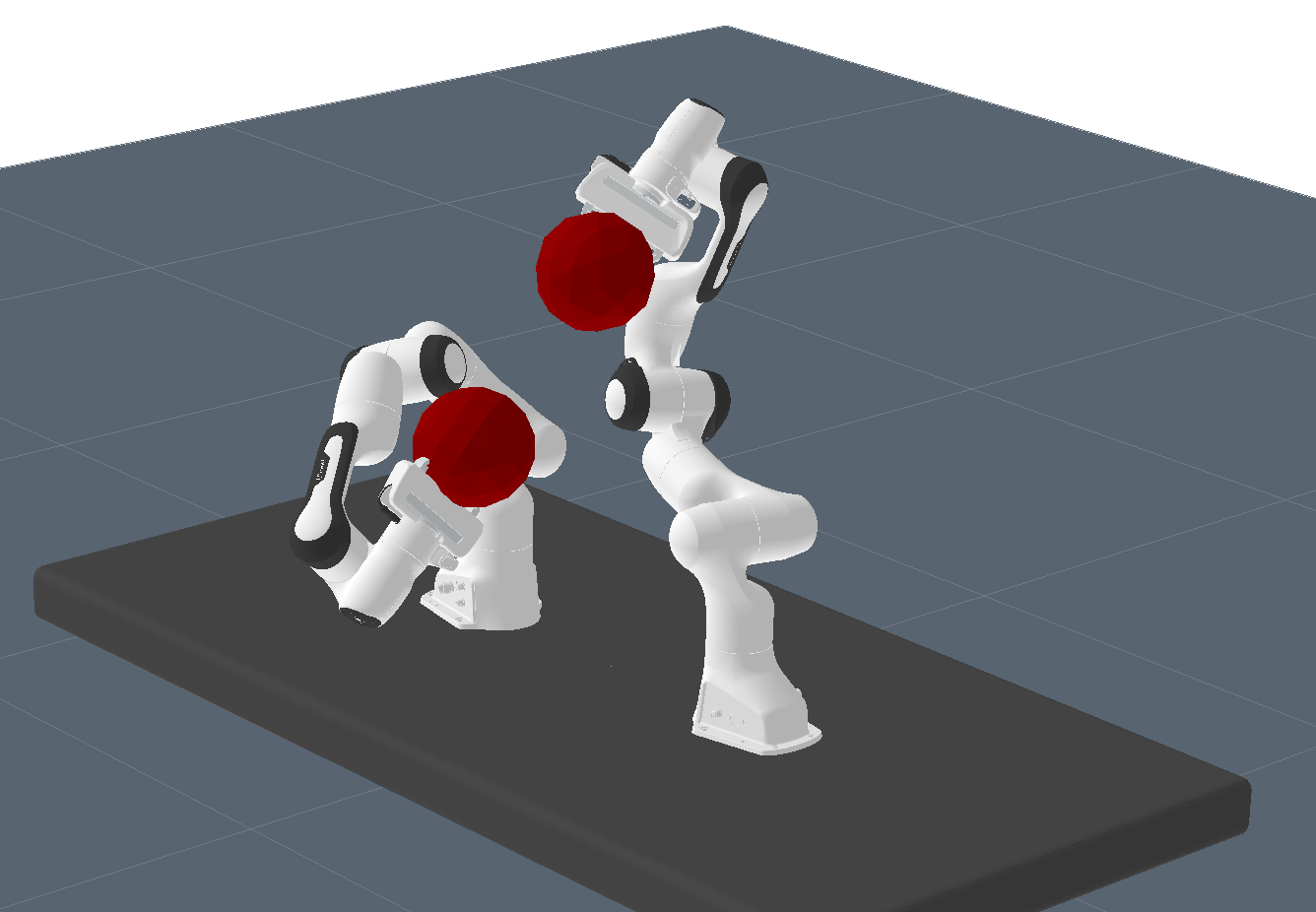}
    \includegraphics[width=0.19\textwidth]{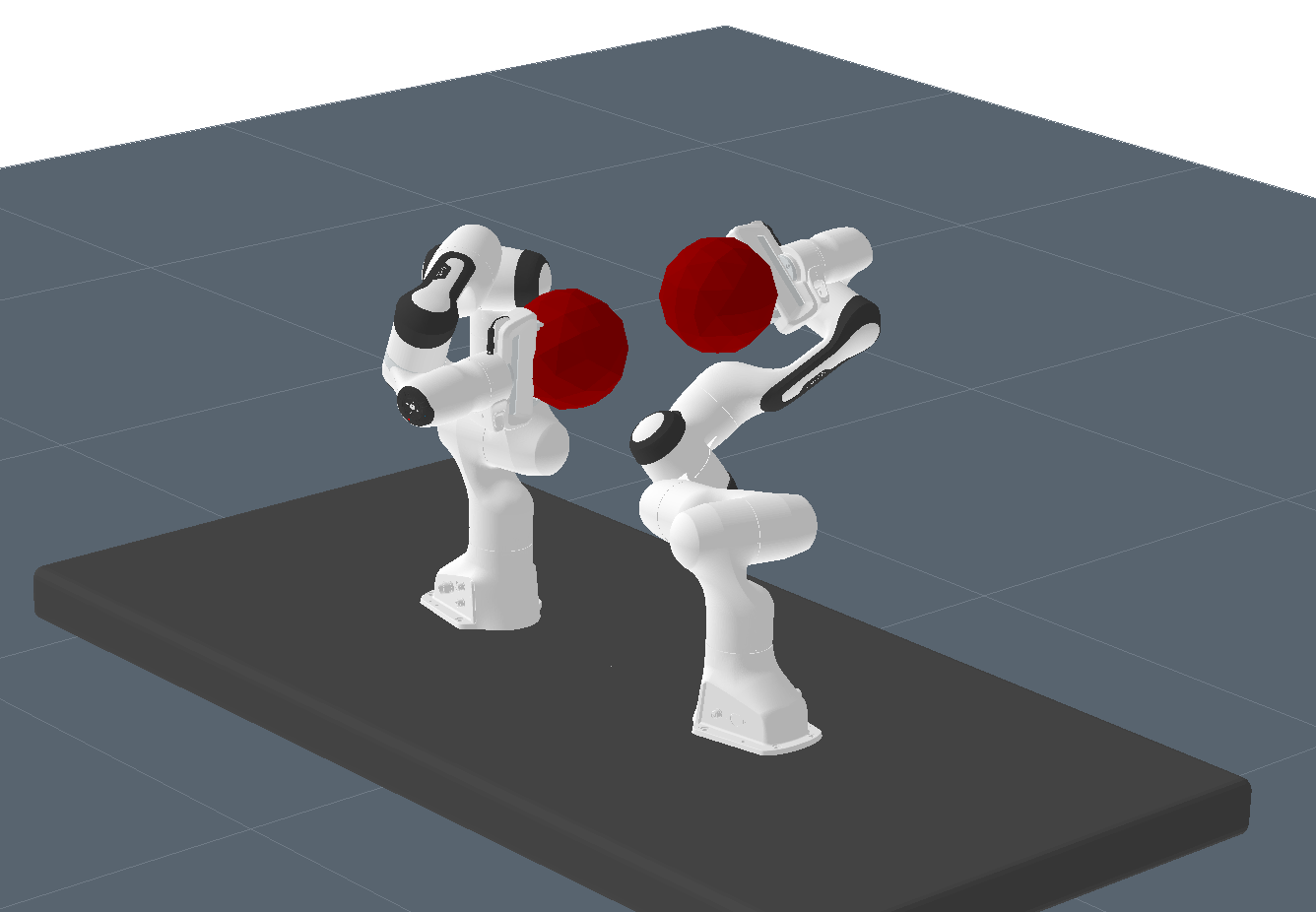}
    \includegraphics[width=0.19\textwidth]{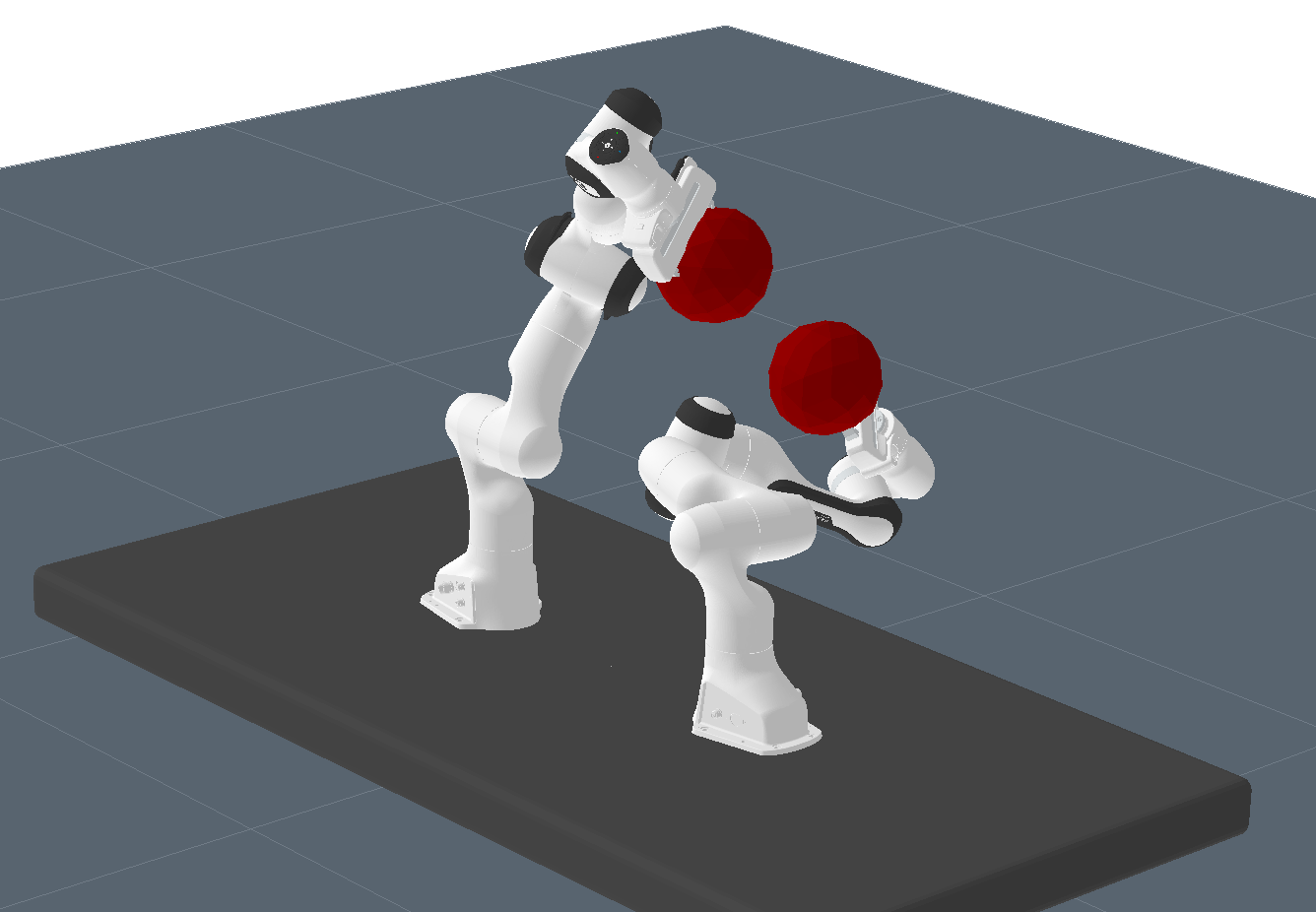}
    \includegraphics[width=0.19\textwidth]{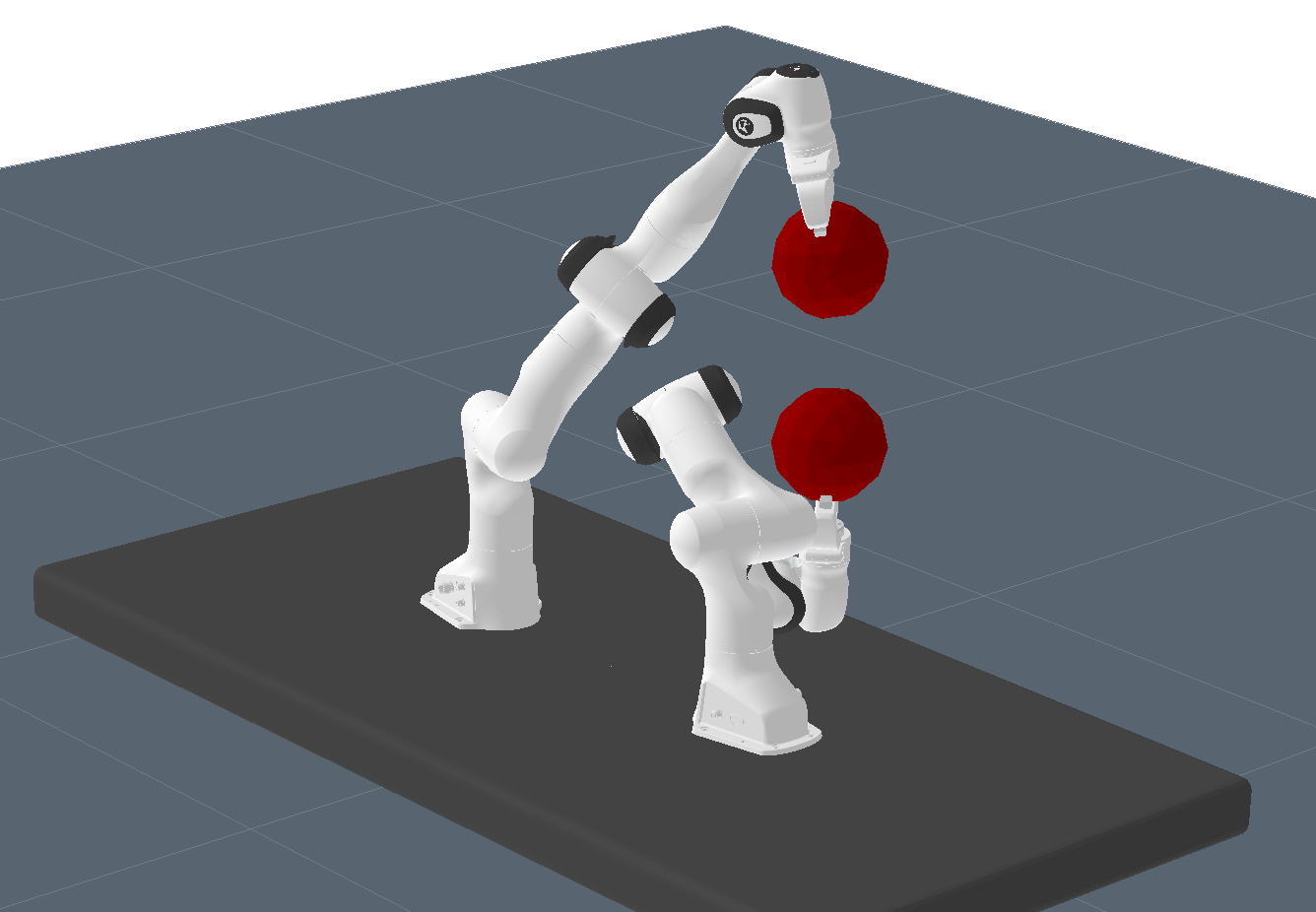}
    \caption{For dexterous bimanual manipulation, it is essential to keep the end effector transformation constant.}
    \label{fig:robot_side_by_side}
\end{figure*}

Second, contact forces are inherently difficult to model and highly discontinuous, making them non-trivial for direct inclusion in optimization problems. 
%Contact forces depend on unpredictable factors such as surface friction, material properties, and environmental uncertainties. 
%As a result, relying solely on precomputed contact forces from optimization can lead to unstable or unrealistic behaviors. Instead, closed-loop feedback systems are required to regulate contact forces in real time, ensuring compliance and adaptability during contact-rich interactions \cite{contactReview}. Simply incorporating contact force constraints into an optimization problem and following the resulting trajectory in an open-loop manner is insufficient for robust manipulation.

The limitations of each paradigm highlight the need for a unified framework that integrates their complementary strengths. Human manipulation serves as a compelling inspiration for such integration. When humans perform tasks, they coordinate their limbs efficiently, leveraging both high-level planning and low-level force regulation. For example, picking up an object involves planning a trajectory to approach the object (optimization) while applying appropriate grip forces to grasp it securely (force control). Similarly, carrying an object requires synchronized arm movements (optimization) while maintaining stable contact forces (force control). To replicate this level of sophistication in robotic systems, clearly both force control and optimization-based motion planning must be utilized (Fig.~\ref{fig:robot_side_by_side}).

We also recognize that not all aspects of a task demand the same type of control. Some subtasks benefit from pure optimization—for instance, navigating through cluttered environments or planning collision-free paths. Others require pure force control, such as maintaining consistent pressure while polishing a surface. Still, others necessitate a hybrid of both, like aligning and inserting a peg into a hole. By decomposing a task into subtasks tailored to specific control modes (i.e., optimization, force control, or both), the specific subtasks can be completed more efficiently and successfully. This sequential switching between optimization and force control ensures versatility and robustness across a wide range of scenarios.

In this work, we propose an optimization-augmented force control framework that bridges the gap between optimization-based motion planning and force control. 
Our framework combines the strengths of both paradigms to enable robots to perform complex manipulation tasks that neither method could robustly achieve alone. Specifically, our contributions include:

\begin{itemize}
\item An optimization-augmented control framework that seamlessly integrates optimization-based planning with force control for complex manipulation tasks. This framework leverages optimization to generate long-horizon, collision-free trajectories while employing force control to regulate contact forces and ensure compliance during contact-rich interactions. The framework supports seamless switching between pure optimization, pure force control, and hybrid modes based on task requirements.

\item Experimental validation across a range of scenarios, showcasing the framework’s applicability in single-arm, bimanual, and multi-arm settings, using our task decomposition strategy that identifies when to use pure optimization, pure force control, or a hybrid of them. Through a series of experiments, we demonstrate the versatility and effectiveness of our framework in solving complex manipulation challenges, including tasks that require precise alignment, compliant interaction, and synchronized multi-arm coordination.
\end{itemize}

This work represents a significant step toward general-purpose robotic manipulation, combining the strengths of optimization and force control to address tasks previously unattainable with either method alone. 
By enabling robots to perform complex, contact-rich tasks with precision and adaptability, our framework paves the way for more advanced robotic systems capable of operating in diverse and dynamic environments.

%\begin{figure}[thpb]
%\label{example}
%    \centering
%    \begin{subfigure}{0.25\textwidth} % Adjust width
%        \centering
%        \includegraphics[width=\linewidth]{blend1.png}
%        \label{fig:subfig1}
%    \end{subfigure}
%    \begin{subfigure}{0.25\textwidth}
%        \centering
%        \includegraphics[width=\linewidth]{blend2.png}
%        \label{fig:subfig2}
%    \end{subfigure}
%    \begin{subfigure}{0.25\textwidth}
%        \centering
%        \includegraphics[width=\linewidth]{blend3.png}
%        \label{fig:subfig3}
%    \end{subfigure}
%    \begin{subfigure}{0.25\textwidth}
%        \centering
%        \includegraphics[width=\linewidth]{blend4.png}
%        \label{fig:subfig4}
%    \end{subfigure}
%    \begin{subfigure}{0.25\textwidth}
%        \centering
%        \includegraphics[width=\linewidth]{blend5.png}
%        \label{fig:subfig4}
%    \end{subfigure}

%    \caption{For dexterous bimanual manipulation, it is essential to keep the end effector transformation constant.}
%    \label{fig:exp1}
%\end{figure}

\section{BACKGROUND AND RELATED WORK}

Force control has been widely used in industrial applications, such as grinding and surface polishing, to achieve compliance during contact-rich interactions \cite{forceControlReview}. Among these methods, hybrid impedance and admittance control is highly adapted for its complementary advantages \cite{hybrid}. Iskandar et al. proposed a hybrid force-impedance control framework to enhance precision in tracking curved surfaces \cite{hybridImpedance}. 
%However, such methods often assume predefined contact states and struggle with dynamic transitions between contact modes. 
Cheng et al. introduced the Contact Mode Guided Motion Planning (CMGMP) framework, which generates hybrid motion plans for quasistatic and quasidynamic manipulation tasks \cite{contactMode}. While their method eliminates the need for pre-defined motion primitives, it relies on instantaneous contact transitions (e.g., contacts appearing or disappearing abruptly), limiting its applicability to real-world scenarios.   

Optimization methods have also been utilized in robotics for trajectory planning \cite{optimTraj}. Recently, Toussaint et al. developed the SECMPC framework, which applies KOMO optimization recursively to solve sequential manipulation tasks \cite{secmpc}. While effective for pick-and-place operations, SECMPC neglects contact force regulation, relying solely on open-loop trajectory execution. This limitation highlights the need for hybrid approaches that integrate force control with optimization. Yan et al. propose an online optimization framework for dynamic dual-arm catching of high-momentum objects \cite{impactAware}. Their method selects optimal contact points via sequential quadratic programming (QP) to distribute impact forces. Chaki et al. used quadratic programming for bimanual manipulation tasks \cite{quadratic}. Their method does not regulate the contact forces explicitly, but they assume that contact forces are linear, i.e. utilize a spring model and demonstrate that some tasks are achievable, like holding three boxes stationary in the air, between two arms, which have a large contact area. In reality, even carrying these boxes while walking necessitates force control because vibration from walking would lead the boxes to drop.   
Cohn et al. emphasized the importance of maintaining a constant transformation between end-effectors in bimanual tasks \cite{tedrakePlanning}. Their approach allows one arm to move freely while the other regulates its pose to satisfy constraints. However, this decoupled strategy often results in infeasible configurations for the follower arm, as it ignores joint limits and global constraints. Joint optimization of both arms is essential for ensuring feasibility in complex tasks. Both force control and optimization have their advantages, and they should be utilized accordingly.

In the recent years, learning-based approaches gained momentum \cite{imitation, diffusionSurvey}.
Advances in learning-based methods have enabled robots to acquire manipulation skills from demonstrations. Zhao et al. developed a bimanual robotic kit that leverages imitation learning with Action Chunking Transformers, enabling robots to generalize from human demonstrations \cite{aloha}. Their framework facilitates efficient learning of dexterous skills but remains limited to demonstrated behaviors.
Chi et al. introduced a diffusion-based policy for robotic control \cite{diffusionPolicy}. Their method computes control inputs over a time horizon but only executes a subset of these before recomputing, making it more adaptable to dynamic environments, a receding-horizon strategy that is widely applied in model predictive controller (MPC) methods \cite{mpc}. This approach bridges aspects of planning and learning, but its performance remains constrained by the quality of the training data.
To address force-adaptive manipulation, Hou et al. tackled the problem of cleaning a vase while another arm stabilizes it \cite{adaptiveCompliancePolicy}. Recognizing that different motion segments require varying levels of compliance, they developed an adaptive compliance policy where force regulation is learned from demonstrations. Similarly, Kamijo et al. learn a variable compliance policy using virtual reality (VR) controllers with haptic feedback \cite{bimanualTeleop}.
Liu et al. argue that although vision-based teleoperation has been the dominant method for data collection in learning-based robotics, force sensing plays a crucial role in manipulation \cite{factr}. Motivated by this observation, Liu et al. developed a bimanual teleoperation hardware system that integrates force sensing alongside vision. This allows operators to both see the scene and physically feel forces, improving demonstration fidelity. Chen et al. employ a similar approach by collecting the force data from the end effector tips \cite{dexForce}. 
While learning-based methods offer promising adaptability, they have fundamental limitations. Unlike optimization-based approaches, they lack explicit guarantees of optimality and cannot generalize well beyond their training distribution. This limits their applicability in unstructured or dynamically changing environments where unforeseen variations may arise.

%\begin{figure}[h]
    %\centering
    %\setlength{\fboxsep}{0pt} % Remove box spacing
    %\includegraphics[width=0.48\textwidth]{blend5.png}
    %\transparent{0.6} % 60% transparent
    %\includegraphics[width=0.48\textwidth]{blend4.png}
    %\transparent{0.4} % 40% transparent
    %\includegraphics[width=0.48\textwidth]{blend3.png}
    %\transparent{0.2} % 20% transparent
    %\includegraphics[width=0.48\textwidth]{blend2.png}
    %\caption{Overlay of robot configurations.}
%\end{figure}

\section{METHODOLOGY}

Our method leverages a multi-modal control framework that seamlessly switches between \textit{i)} pure optimization, \textit{ii)}~pure force control, and \textit{iii)} hybrid optimization with force control to accomplish complex robotic manipulation tasks. Given a high-level manipulation task, it is first decomposed into subtasks, each assigned to one of the three controllers based on its requirements. By combining these subtasks sequentially, the overall task is successfully executed. 

Next, we present our task decomposition strategy and the controller design, including mathematical formulations and implementation details. 
% A given robotic manipulation task is decomposed into subtasks categorized into three modes:

\subsection{Task Decomposition}

The foundation of our approach lies in decomposing a given manipulation task into subtasks, each tailored to one of three control modes: pure optimization, pure force control, or hybrid optimization and force control. This decomposition ensures that the most appropriate controller is selected for each subtask, maximizing efficiency and robustness. In the current implementation, task decomposition is performed manually, although automating this process is a desirable future direction. 
A given robotic manipulation task is decomposed into subtasks, which are categorized into three modes:

\begin{enumerate}

  \item \textit{Pure Optimization Subtasks}: They are employed when contact forces and object inertia are not critical to the task. Examples include pick-and-place operations, iterative pushing of lightweight objects (e.g., a small box) toward a target, or free-space motion where collisions and physical interactions are minimal. In such cases, the robot relies solely on motion planning to achieve the desired trajectory without explicitly regulating contact forces.
  
  \item \textit{Pure Force Control Subtasks}: These subtasks are essential when precise force regulation is required, and the end-effector orientation is less critical. For instance, applying a consistent force while moving along a surface or rotating an object (e.g., making a box upright) falls under this category. Pure force control excels in high-frequency, short-range manipulations where compliance and adaptability are paramount.
  
  \item \textit{Hybrid Optimization and Force Control Subtasks}: These subtasks arise when both force regulation and precise pose control are necessary. A prime example is bimanual manipulation, such as carrying a box using two robot arms. Here, the arms must maintain stable grasps by controlling both their positions and orientations while applying synchronized forces to the object. Optimization-augmented force control is particularly advantageous for multi-arm coordination, where the interplay between arms and the manipulated object introduces additional constraints.
\end{enumerate}

\subsubsection*{Selection of Contact Points}

The selection of contact points significantly influences force distribution and stability. In general, we choose antipodal points that pass through the object's geometric center, ensuring that:

\begin{itemize}

\item No unwanted torques are generated.

\item The grasp remains firm and stable throughout the manipulation process.

\end{itemize}

However, in some cases (such as object rotation) a sliding contact point is preferable, allowing controlled motion around a pivot. The selection of static vs. sliding contact points is determined based on the task requirements.

\subsection{Controller Design}

Our control framework is based on \textit{inverse dynamics control} combined with task-specific control laws for optimization and force regulation. The robot arm dynamics are governed by \cite{robotics}:
\begin{equation}
M(\mathbf{q}) \mathbf{\ddot{q}} + C(\mathbf{q}, \mathbf{\dot{q}}) \mathbf{\dot{q}} + G(\mathbf{q}) = \mathbf{\tau}
\end{equation}
where:
\begin{itemize}
\item $M(\mathbf{q})$ is the mass/inertia matrix.
\item $C(\mathbf{q}, \mathbf{\dot{q}})$ is the Coriolis and centrifugal terms.
\item $G(\mathbf{q})$ is the gravity vector.
\item $\mathbf{\tau}$  is the joint torque vector.
\end{itemize}

To track desired joint trajectories, we use an inverse dynamics control law
\begin{equation}
\mathbf{\tau} = M(\mathbf{q}) \mathbf{\ddot{q}}_{\text{des}} + C(\mathbf{q}, \mathbf{\dot{q}}) \mathbf{\dot{q}}_{\text{des}} + G(\mathbf{q}) + \mathbf{\tau}_{\text{cmd}}
\end{equation}
where we use a PID controller for joint position tracking
\begin{equation}
\begin{aligned}
\label{eq:pid}    
\mathbf{\tau}_{\text{cmd}} = \mathbf{\tau}_{\text{optimization}} &= K_p (\mathbf{q}_{\text{des}} - \mathbf{q}) + K_d (\mathbf{\dot{q}}_{\text{des}} - \mathbf{\dot{q}}) \\
&\quad + K_i \int (\mathbf{q}_{\text{des}} - \mathbf{q}) dt
\end{aligned}
\end{equation}
This allows smooth trajectory tracking, ensuring accurate execution of optimized and force-regulated motions.

\subsubsection{Optimization-Based Control (KOMO)}

We employ KOMO (K-Order Markov Constrained Optimization) for trajectory planning \cite{komo}:
\begin{equation}
\begin{aligned}
\label{eq:komo}
\min_{q_{0:T}} \quad & \sum_{t=0}^{T} f_t(q_{t-k:t})^T f_t(q_{t-k:t}) \\
\text{s.t.} \quad & g_t(q_{t-k:t}) \leq 0, \quad h_t(q_{t-k:t}) = 0
\end{aligned}
\end{equation}
where:
\begin{itemize}
\item $f_t(q_{t-k:t})$ is the cost function penalizing deviation from the desired path,

\item $g_t(q_{t-k:t})$ is the inequality constraints (e.g., collision avoidance, joint limits),

\item $h_t(q_{t-k:t})$ is the equality constraints (e.g., end-effector goal position).
\end{itemize}

Once a feasible trajectory $q_{\text{ref}}$ is obtained from KOMO, it is tracked using a PID controller
\begin{equation}
\begin{aligned}
\label{eq:komo_pid}
\tau_{\text{optimization}} &= K_p (q_{\text{ref}} - q) + K_d (\dot{q}_{\text{ref}} - \dot{q}) \\
&\quad + K_i \int (q_{\text{ref}} - q) dt    
\end{aligned}
\end{equation}

%In order to operate efficiently, we check path feasibility dynamically. If the path remains collision-free, we execute it without re-solving KOMO. If infeasible, we recompute the trajectory.

\subsubsection{Force Control}

Force control is used when interaction forces are critical, and it consists of two components:

% \begin{enumerate}[(a)]
\paragraph{Contact Force Regulation (PI Controller)}
%   % \item \textit{Contact Force Regulation (PI Controller)}: 
Contact forces are highly discontinuous and must be regulated frequently \cite{contactReview}. We use a PI controller
\begin{equation}
F_{\text{contact}} = K_p (F_{\text{target}} - F_{\text{actual}}) + K_i \int (F_{\text{target}} - F_{\text{actual}}) dt
\end{equation}
\begin{equation}
\tau_{\text{contact}} = J^T F
\end{equation}
The PI controller adapts quickly to changes in contact conditions, ensuring stable and compliant interactions.

\paragraph{Position Control (Cartesian Impedance Control)}
%   % \item \textit{Position Control (Cartesian Impedance Control)}: 
For precise free-space motion, we employ a Cartesian Impedance Controller , which provides precise position and orientation control while maintaining compliance \cite{impedance}:
\begin{equation}
F_{\text{position}} = K_p (x_{\text{des}} - x) + D_p (\dot{x}_{\text{des}} - \dot{x})
\end{equation}
\begin{equation}
\tau_{\text{position}} = J^T F
\end{equation}

This approach is particularly effective for short-range manipulations where the end-effector orientation is less critical. Additionally, force control serves as a fallback mechanism when the optimization-based planner fails to find a feasible solution within the given time constraints. For instance, if KOMO cannot resolve a synchronous motion for two arms moving 30 cm upward, we switch to pure force control, which can handle such short-distance motions without requiring precise orientation control.

Note that, when we control both position and contact force, using force control approach, the total torque to be applied is the summation of both torques
\begin{equation}
\tau_{\text{cmd}} = \tau_{\text{contact}} + \tau_{\text{position}}
\end{equation}
% \end{enumerate}

\subsubsection{Optimization-Augmented Force Control}

The hybrid optimization and force control mode integrates the strengths of both approaches, enabling simultaneous regulation of contact forces and precise pose control. This mode is particularly suited for tasks involving multi-arm manipulation, such as carrying a box with two arms (Fig. \ref{fig:exp2}). In this mode, both trajectory tracking (KOMO) and force regulation (PI controller) are used.

At first, we formulate the whole optimization as a KOMO problem with an added contact force objective
\begin{equation}
\min_{q_{0:T}} \sum_{t=0}^{T} f_t(q_{t-k:t})^T f_t(q_{t-k:t}) + \lambda_{\text{force}} ||F_t - F_{\text{target}}||^2
\end{equation}
However, contact forces are difficult to model and highly discontinuous, making their prediction unreliable. Even when they are predictable, contact forces must be frequently regulated, because a sharp rise/drop in them could result in the failure of the task \cite{contactReview}. Therefore, we control contact forces separately using a PI controller while tracking the planned trajectory, and remove the contact force constraint from the KOMO problem, returning back to its original version (eq.~\ref{eq:komo}).
% \begin{equation}
% \min_{q_{0:T}} \sum_{t=0}^{T} f_t(q_{t-k:t})^T f_t(q_{t-k:t})
% \end{equation}
In this case, the commanded torque can be calculated by adding the torques coming from the PID controller of the optimization and the PI controller of the force regulation terms
\begin{equation}
\label{eq:t_cmd}
  \tau_{\text{cmd}} = \tau_{\text{optimization}} + \tau_{\text{contact}}
\end{equation}

\paragraph*{Practical Stability of Optimization-Augmented Force Control} At the core of our method lies the hybrid controller of optimization with contact force regulation. As seen from Equation~\ref{eq:t_cmd}, the total torque $\tau_{\text{cmd}}$ consists of the torques applied for motion planning $\tau_{\text{optimization}}$, and contact force regulation $\tau_{\text{contact}}$. Our purpose here is to control the pose (position and orientation) of the robot arm along with the force applied to the object. Due to the highly coupled dynamics of the robot arm, these two torques interfere with one another. Considering there will possibly exist other robot arms, and they will be coupled through the object, it will be even more difficult to track the optimization path while applying the target force. In order to design the method for better stability, we make the following assumption: Direction of motion of the robot (obtained from optimization) is orthogonal to the contact force direction, hence their effect on each other is minimal, and small errors can be stabilized by finely-tuned integral terms from both controllers. Based on this assumption, we make the following design choices:

\begin{itemize}
\item Contact points are chosen antipodal and the force vectors also pass from the geometric center of the object, for a solid grasp and minimum undesired moment.
\item Robot arms are commanded to have a desirable pose prior to contact, which allows the next waypoints to be achievable.
\item Contact force controllers are started once all robots get into contact, ensuring robots are in close proximity of their initial target pose.
\end{itemize}

%\begin{itemize}
%\item Contact points are chosen antipodal and the force vectors also pass from the geometric center of the object, for a solid grasp and minimum undesired moment.
%\item Robot arms commanded to have a desirable pose prior to contact, which allows the next waypoints to be achievable.
%\item Contact force controllers are started once all robots get into contact, ensuring robots are in close proximity of their initial target pose.
%\item Direction of contact force and motion are orthogonal, hence their torques have minimum effect on each other. Ideally, contact force does not cause motion, and motion planning does not affect contact force.
%\item Due to uncertainties, they affect each other, but both contact force and optimization term have an integral component. Thus, if tuned appropriately, they will correct the errors good enough to allow the task to be completed.
%\end{itemize}

\begin{figure*}[h]
    \centering
    \includegraphics[width=1\textwidth]{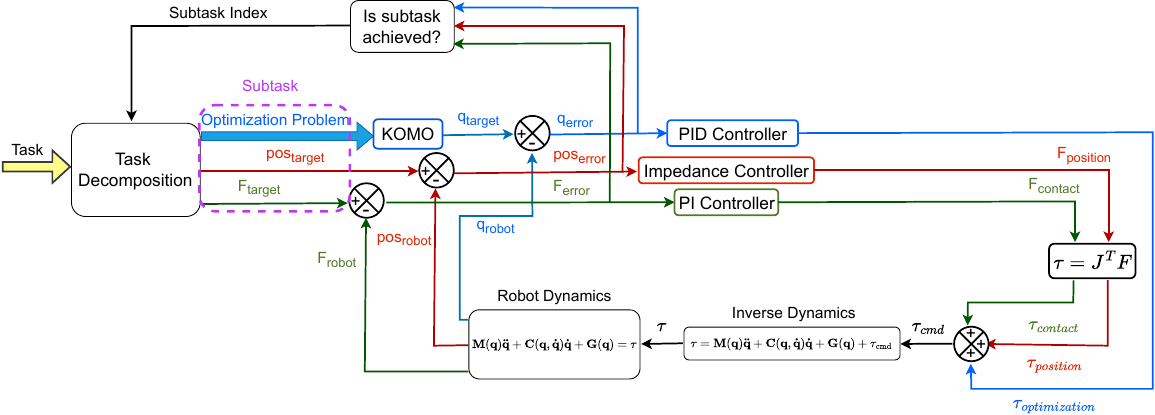} % Adjust width
    \caption{System diagram.}
    \label{fig:system_diagram}
\end{figure*}

Pseudo code for the algorithm is provided (Algorithm \ref{alg:hybrid_control}), and its flowchart is given in Fig. \ref{fig:system_diagram}. In the next section, we will analyze test cases and show that this practical stability intuition indeed holds true.

\begin{algorithm}[h]
%\small
\caption{Optimization-Augmented Force Control}
\label{alg:hybrid_control}
\begin{algorithmic}[1]
\Require Target configuration $\mathbf{q}_{\text{target}}$, Target force $\mathbf{F}_{\text{target}}$, Target position in task space $\mathbf{pos}_{\text{target}}$
\Require Robot state: $\mathbf{q}$, $\mathbf{\dot{q}}$, $\mathbf{F}$, $\mathbf{pos}$
\Require Robot dynamics: Mass matrix $\mathbf{M}(\mathbf{q})$, Coriolis matrix $\mathbf{C}(\mathbf{q}, \mathbf{\dot{q}})$, Gravity vector $\mathbf{G}(\mathbf{q})$
\Require Controller gains: $K_p$, $K_d$, $K_i$ (PID), $K_p^F$, $K_i^F$ (PI for force), $K_p^x$, $K_d^x$ (Impedance)
\Require Threshold $\epsilon$, $\epsilon_{pos}$

\While{$(\|\mathbf{q}_{\text{target}} - \mathbf{q}\| > \epsilon \lor \|\mathbf{pos}_{\text{target}} - \mathbf{pos}\| > \epsilon_{pos})$}

    \State \textbf{// Compute Optimization-Based Control (PID)}
    \State $\mathbf{\tau}_{\text{optimization}} = K_p (\mathbf{q}_{\text{target}} - \mathbf{q}) + K_d (\mathbf{\dot{q}}_{\text{target}} - \mathbf{\dot{q}}) + K_i \int (\mathbf{q}_{\text{target}} - \mathbf{q}) dt$

    \State \textbf{// Compute Force Control (PI)}
    \State $\mathbf{F}_{\text{error}} = \mathbf{F}_{\text{target}} - \mathbf{F}$
    \State $\mathbf{F}_{\text{cmd}} = K_p^F \mathbf{F}_{\text{error}} + K_i^F \int \mathbf{F}_{\text{error}} dt$
    \State $\mathbf{\tau}_{\text{contact}} = \mathbf{J}^T \mathbf{F}_{\text{cmd}}$

    \State \textbf{// Compute Cartesien Impedance Control }
    \State $\mathbf{x}_{\text{error}} = \mathbf{pos}_{\text{target}} - \mathbf{pos}$
    \State $\mathbf{F}_{\text{impedance}} = K_p^x \mathbf{x}_{\text{error}} + K_d^x (\mathbf{\dot{pos}}_{\text{target}} - \mathbf{\dot{pos}})$
    \State $\mathbf{\tau}_{\text{position}} = \mathbf{J}^T \mathbf{F}_{\text{impedance}}$

    \State \textbf{// Compute Final Commanded Torque}
    \State $\mathbf{\tau}_{\text{cmd}} = \mathbf{\tau}_{\text{optimization}} + \mathbf{\tau}_{\text{contact}} + \mathbf{\tau}_{\text{position}}$

    \State \textbf{// Apply Inverse Dynamics Compensation}
    \State $\mathbf{\tau} = \mathbf{M}(\mathbf{q}) \mathbf{\ddot{q}} + \mathbf{C}(\mathbf{q}, \mathbf{\dot{q}}) \mathbf{\dot{q}} + \mathbf{G}(\mathbf{q}) + \mathbf{\tau}_{\text{cmd}}$

    \State \textbf{// Apply Torque to Robot Actuators}
    \State Send $\mathbf{\tau}$ to robot

    \State \textbf{// Update Robot State}
    \State Read $\mathbf{q}, \mathbf{\dot{q}}, \mathbf{F}, \mathbf{pos}$
\EndWhile
\end{algorithmic}
\end{algorithm}

%\begin{figure}[h]
%\centering
%\includegraphics[width=0.48\textwidth]{controlDiagram.pdf}
%\caption{System diagram.}
%\end{figure}

\section{EXPERIMENTS \& EVALUATIONS}

To validate the effectiveness of the proposed hybrid control framework, we conduct a series of simulation experiments with varying configurations and task requirements (see supplementary video). The experiments are performed in the RAI simulation environment, which is chosen for its seamless integration with the KOMO optimization framework \cite{toussaintrai}. 

A 7-DOF Panda robotic arm is used in the experiments, with its end effector replaced by a spherical object to ensure point contact. The simulation physics is handled by the NVIDIA PhysX engine, which we utilize for both robot control and contact force measurement.

\paragraph{Single-Arm Box Relocation}

In this experiment, a box with dimensions $15 \times 15 \times 30$ cm is manipulated in two sequential stages: first, it is rotated to change its orientation, and then it is transported onto a larger platform with a height of $50$ cm (Fig.~\ref{fig:exp1}). The task consists of the following subtasks:

\begin{enumerate}
\item Approach the box $\rightarrow$ \textit{Optimization}
\item Rotate the box $\rightarrow$ \textit{Force Control}
\item Reposition the end effector to grasp the box for transportation $\rightarrow$ \textit{Optimization}
\item Carry the box onto the platform while maintaining end-effector orientation perpendicular to the moving direction $\rightarrow$ \textit{Optimization-Augmented Force Control}
\end{enumerate}

%Maintaining a consistent end-effector orientation during transportation ensures a stable contact force throughout the motion. 
As the box reaches the top of the platform, it undergoes minor rotational adjustments. Keeping the end-effector orientation fixed prevents excessive slipping and ensures that the contact point with the box remains in close proximity. 

%Although the platform acts as a support surface, the robot still regulates the applied force to avoid excessive impact while transitioning onto the platform. This experiment highlights the ability of our framework to dynamically switch between optimization-based motion planning and force control to achieve precise and stable object manipulation.

%\begin{figure}[thpb]
%\label{task1}
%\vspace{-4mm}
%    \centering
%    \begin{subfigure}{0.238\textwidth} % Adjust width
%        \centering
%        \includegraphics[width=\linewidth]{2a0.png}
%        \caption{approach the box}
%        \label{fig:subfig1}
%    \end{subfigure}
%    \begin{subfigure}{0.238\textwidth}
%        \centering
%        \includegraphics[width=\linewidth]{2a.png}
%        \caption{rotate it}
%        \label{fig:subfig2}
%    \end{subfigure}
    
    % \vspace{0.2cm} % Add vertical space between rows

%    \begin{subfigure}{0.238\textwidth}
%    \vspace{-6mm}
%        \centering
%        \includegraphics[width=\linewidth]{2b.png}
%        \caption{carry the box upward}
%        \label{fig:subfig3}
%    \end{subfigure}
%    \begin{subfigure}{0.238\textwidth}
%        \centering
%        \includegraphics[width=\linewidth]{2d.png}
%        \caption{bring it onto the platform}
%        \label{fig:subfig4}
%    \end{subfigure}

%    \caption{Robot approaches the box, rotates it, then carries it onto the platform while keeping its end effector perpendicular to the motion direction.}
%    \label{fig:exp1}
%\end{figure}

\paragraph{Bimanual Box Transportation} \label{boxTransportation}

\begin{figure}[thpb]
\label{task2}
    \centering
    \begin{subfigure}{0.238\textwidth} % Adjust width
        \centering
        \includegraphics[width=\linewidth]{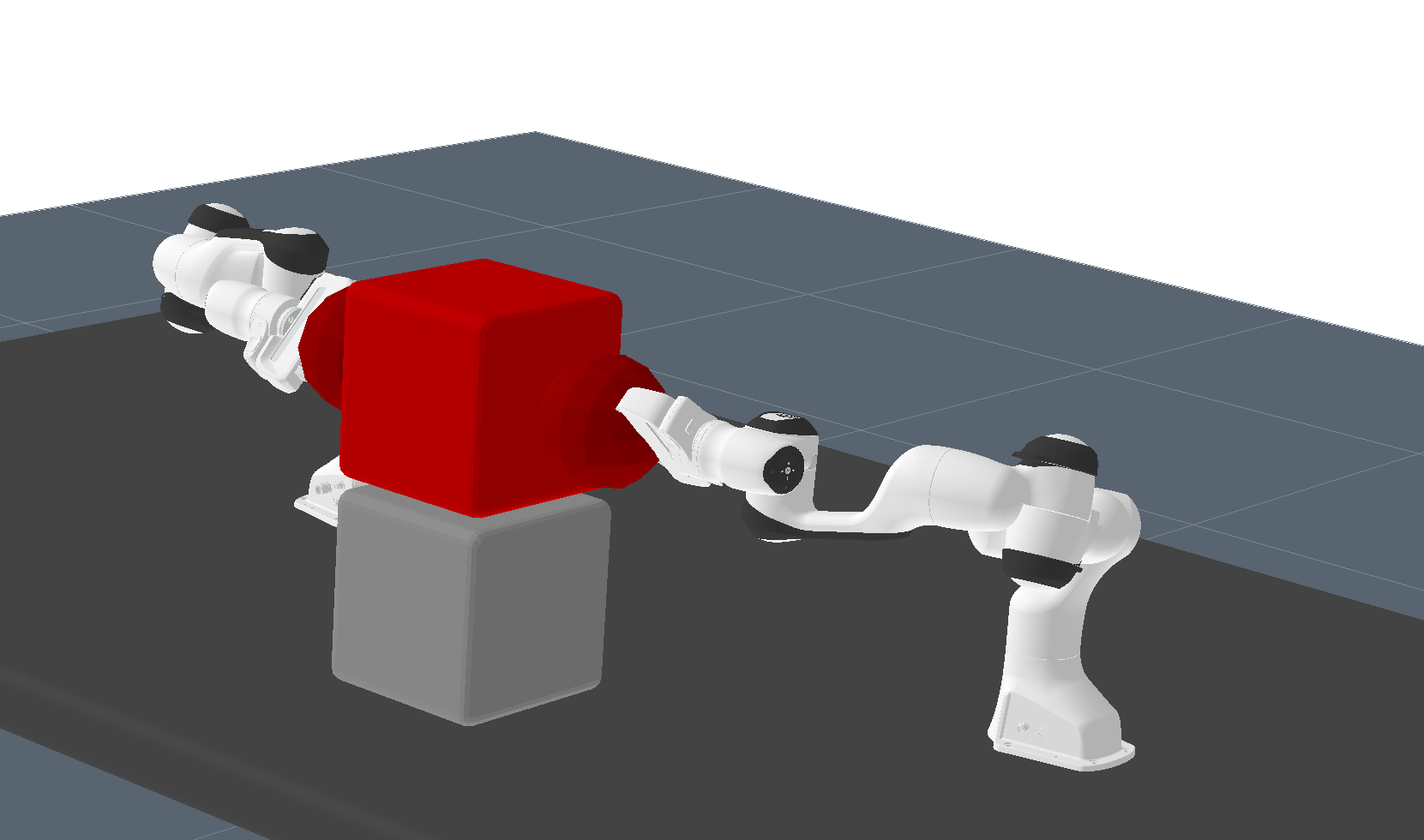}
        \caption{grasp the box on top}
        \label{fig:subfig100}
    \end{subfigure}
    \begin{subfigure}{0.238\textwidth}
        \centering
        \includegraphics[width=\linewidth]{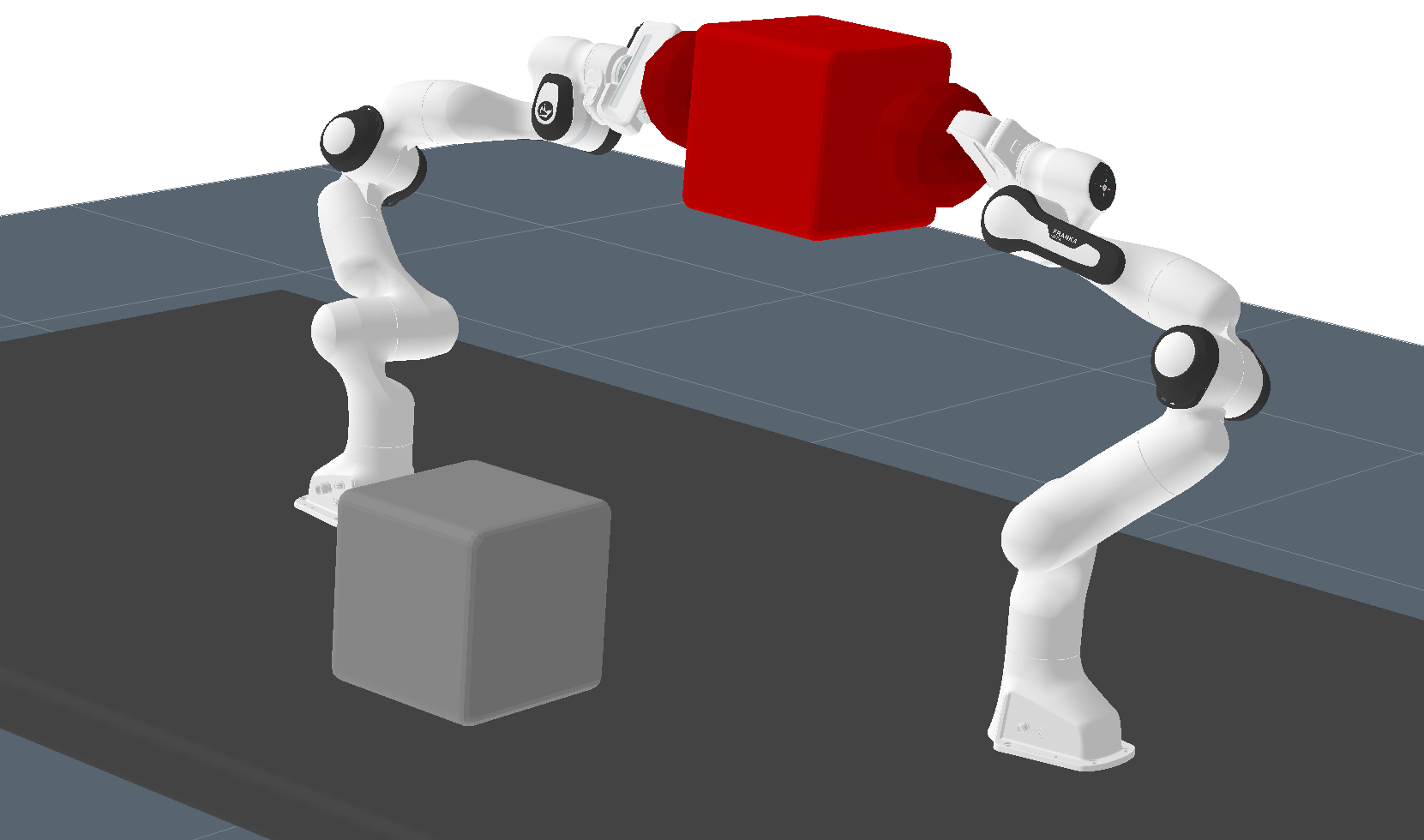}
        \caption{bring it 1m further}
        \label{fig:subfig200}
    \end{subfigure}
    
    \vspace{0.2cm} % Add vertical space between rows

    \begin{subfigure}{0.238\textwidth}
        \centering
        \includegraphics[width=\linewidth]{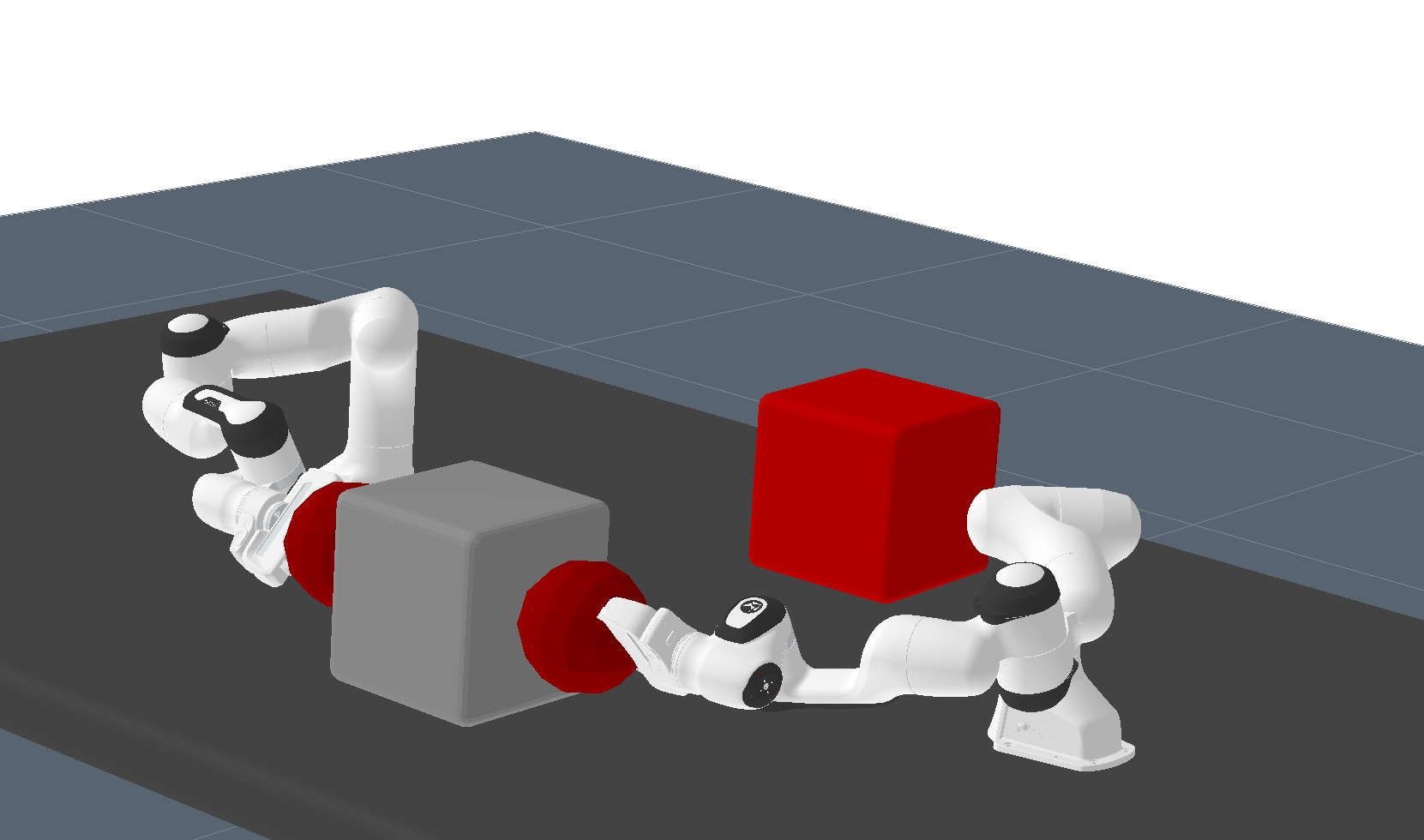}
        \caption{move to the second box}
        \label{fig:subfig300}
    \end{subfigure}
    \begin{subfigure}{0.238\textwidth}
        \centering
        \includegraphics[width=\linewidth]{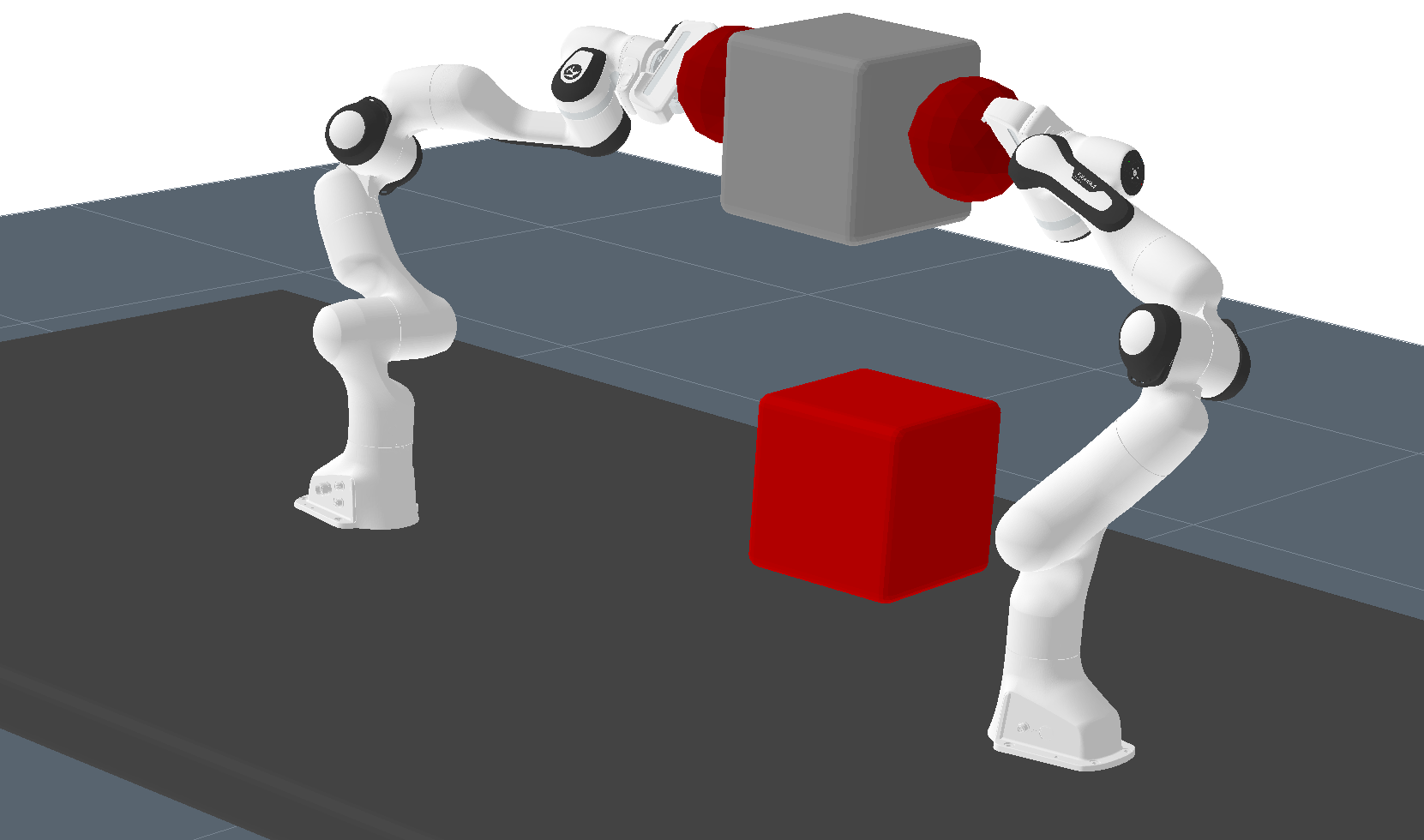}
        \caption{put it onto the first box}
        \label{fig:subfig400}
    \end{subfigure}

    \caption{Robots transport two boxes for 1 meter, by grasping from antipodal points. This kind of extended motions require precise end effector coordination, which is why optimization is essential.}
    \label{fig:exp2}
\end{figure}

Two cubes whose edges are $30 cm$ are carried $1$ meter away by two robot arms that are $1.4 m$  apart (Fig.~\ref{fig:exp2}). This kind of motion requires precise end effector control for both position and orientation, coordinated synchronously for both arms; which is achieved by combined optimization and contact force control. This task reaffirms that our method allows the robots to work effectively in their operational space range.

\paragraph{Bimanual Rotation \& Peg-in-Hole}

\begin{figure}[thpb]
\label{task3}
    \centering
    \begin{subfigure}{0.238\textwidth} % Adjust width
        \centering
        \includegraphics[width=\linewidth]{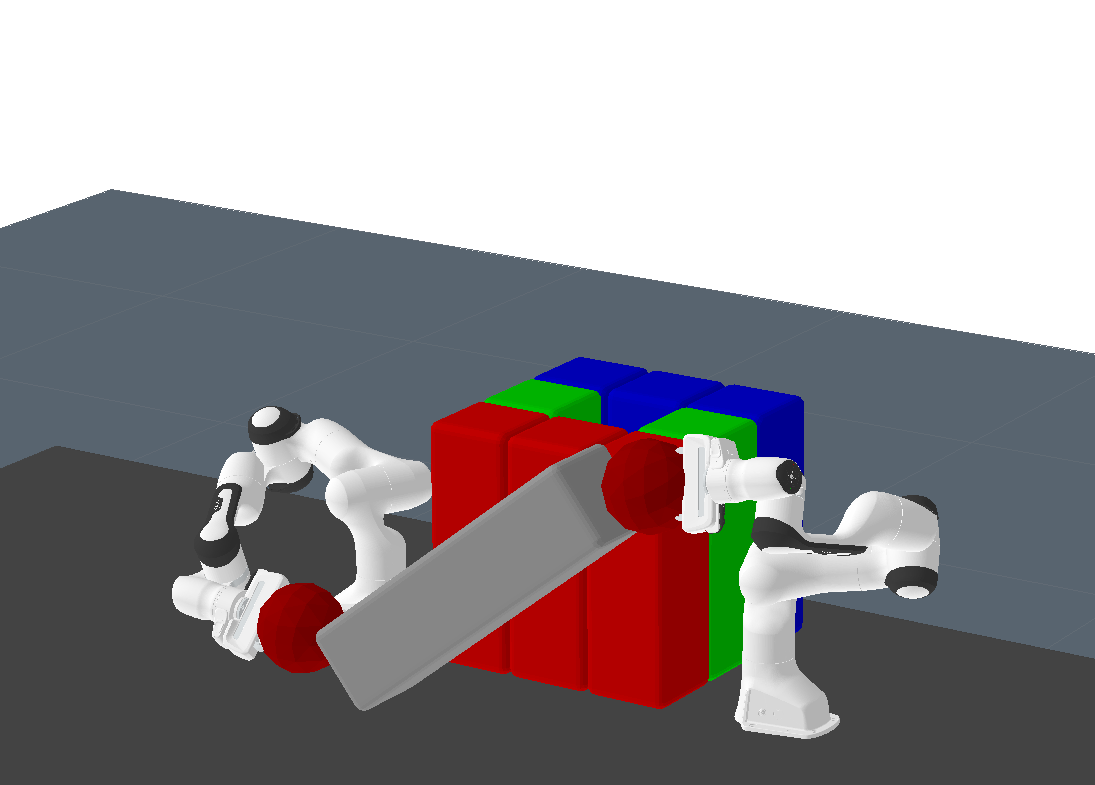}
        \caption{rotate the object}
        \label{fig:subfig1000}
    \end{subfigure}
    \begin{subfigure}{0.238\textwidth}
        \centering
        \includegraphics[width=\linewidth]{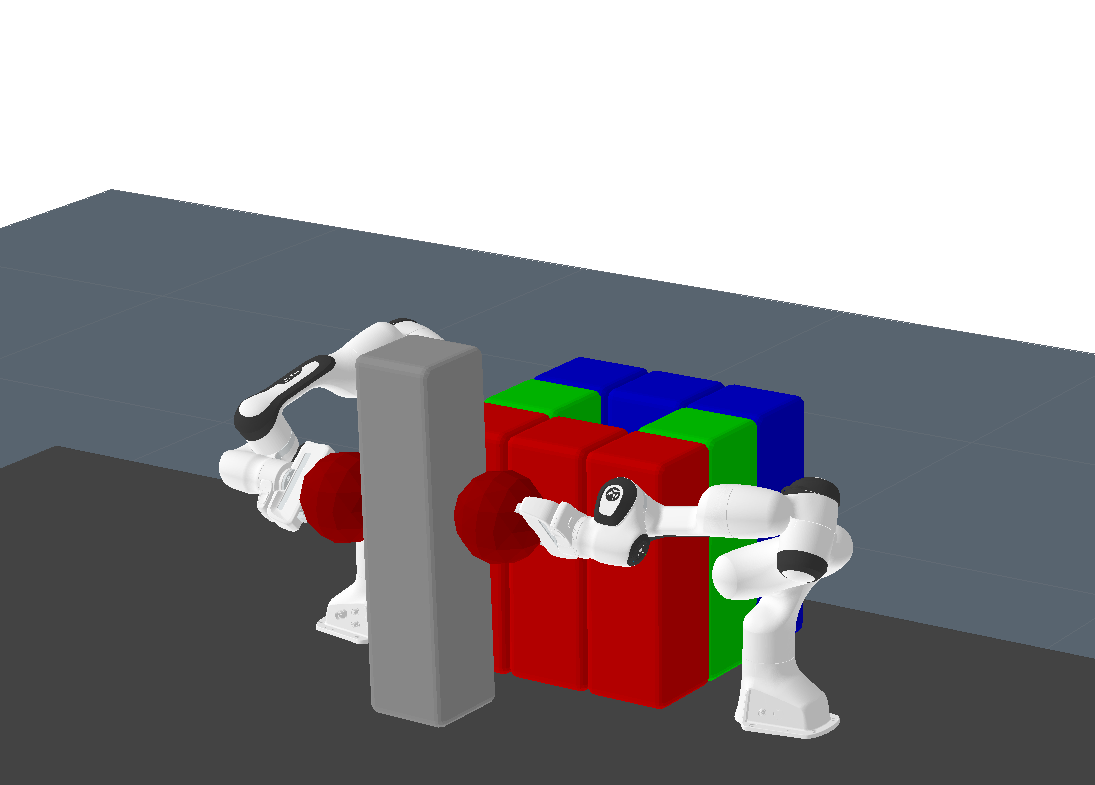}
        \caption{grasp from antipodal points}
        \label{fig:subfig2000}
    \end{subfigure}
    
    \vspace{0.2cm} % Add vertical space between rows

    \begin{subfigure}{0.238\textwidth}
        \centering
        \includegraphics[width=\linewidth]{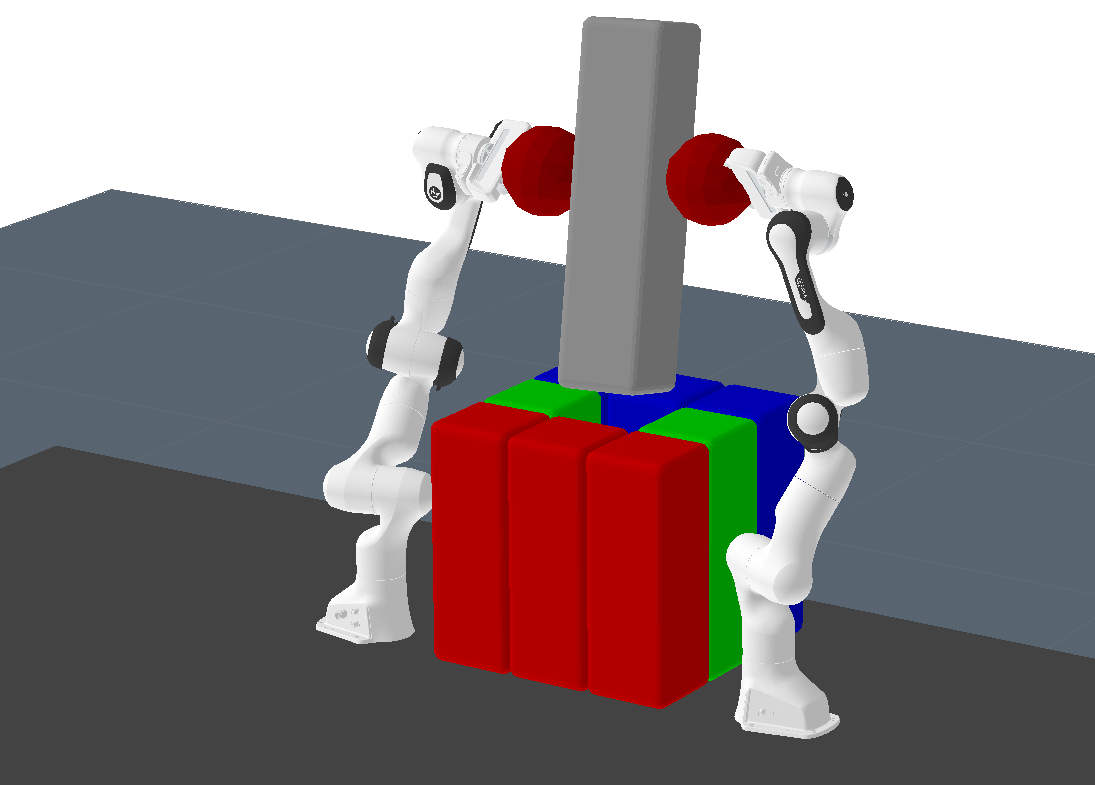}
        \caption{position it above the hole}
        \label{fig:subfig3000}
    \end{subfigure}
    \begin{subfigure}{0.238\textwidth}
        \centering
        \includegraphics[width=\linewidth]{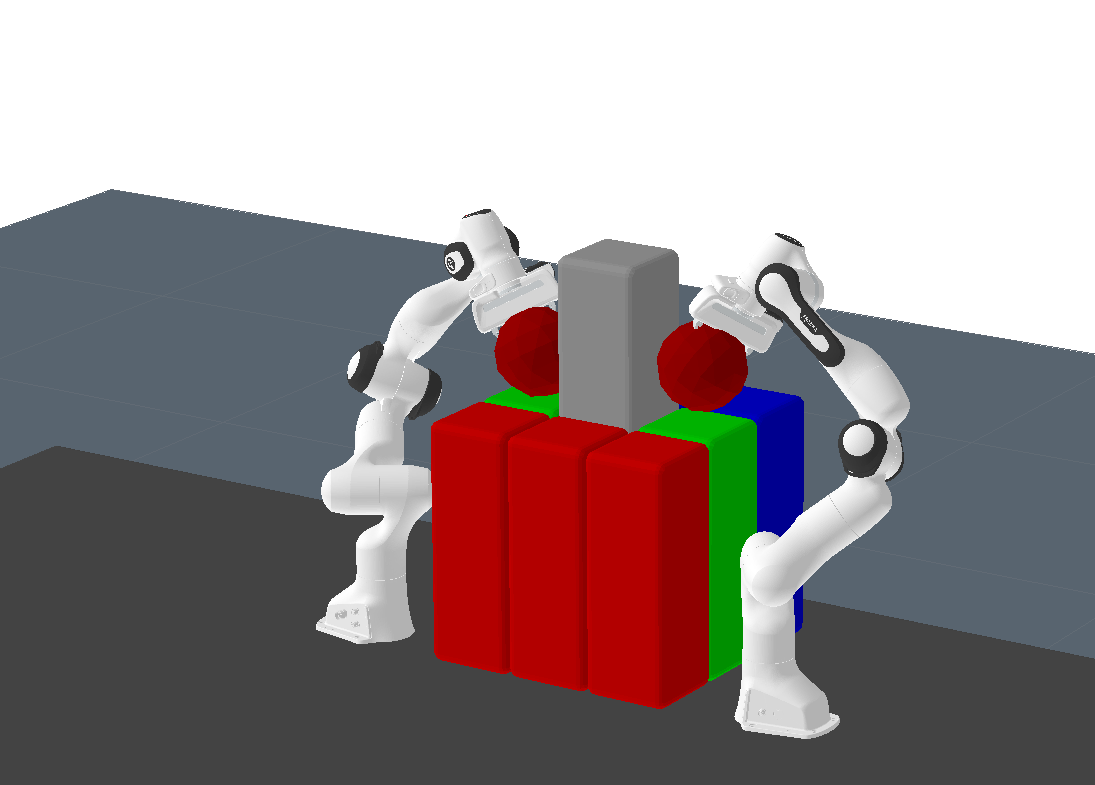}
        \caption{peg into the hole}
        \label{fig:subfig4000}
    \end{subfigure}

    \caption{Two robot arms first rotate an object, then position above the hole, and finally peg the box into the whole. The robots work in the limits of their operational space reach.}
    \label{fig:exp3}
\end{figure}

An object of size $20x20x80 cm$ is rotated and put into the boxes with same profile and a height of $55 cm$ (Fig.~\ref{fig:exp3}).
%To tackle this task, one needs to switch between optimization-based force control and hybrid force control sequentially. 
Rotation of the box requires hybrid force control, then grasping it from antipodal points and carrying it to the pre-insertion position necessitates synchronous motion between the arms, hence combined optimization and contact force control. 
Finally, during insertion, hybrid force control is used once again. 
During this task, the robots approach their operational range, while keeping the desired relative orientation, showing the strength of optimization-augmented force control again.

\paragraph{Multi-Arm Reorientation}

\begin{figure}[thpb]
\label{task4}
    \centering
    \begin{subfigure}{0.238\textwidth} % Adjust width
        \centering
        \includegraphics[width=\linewidth]{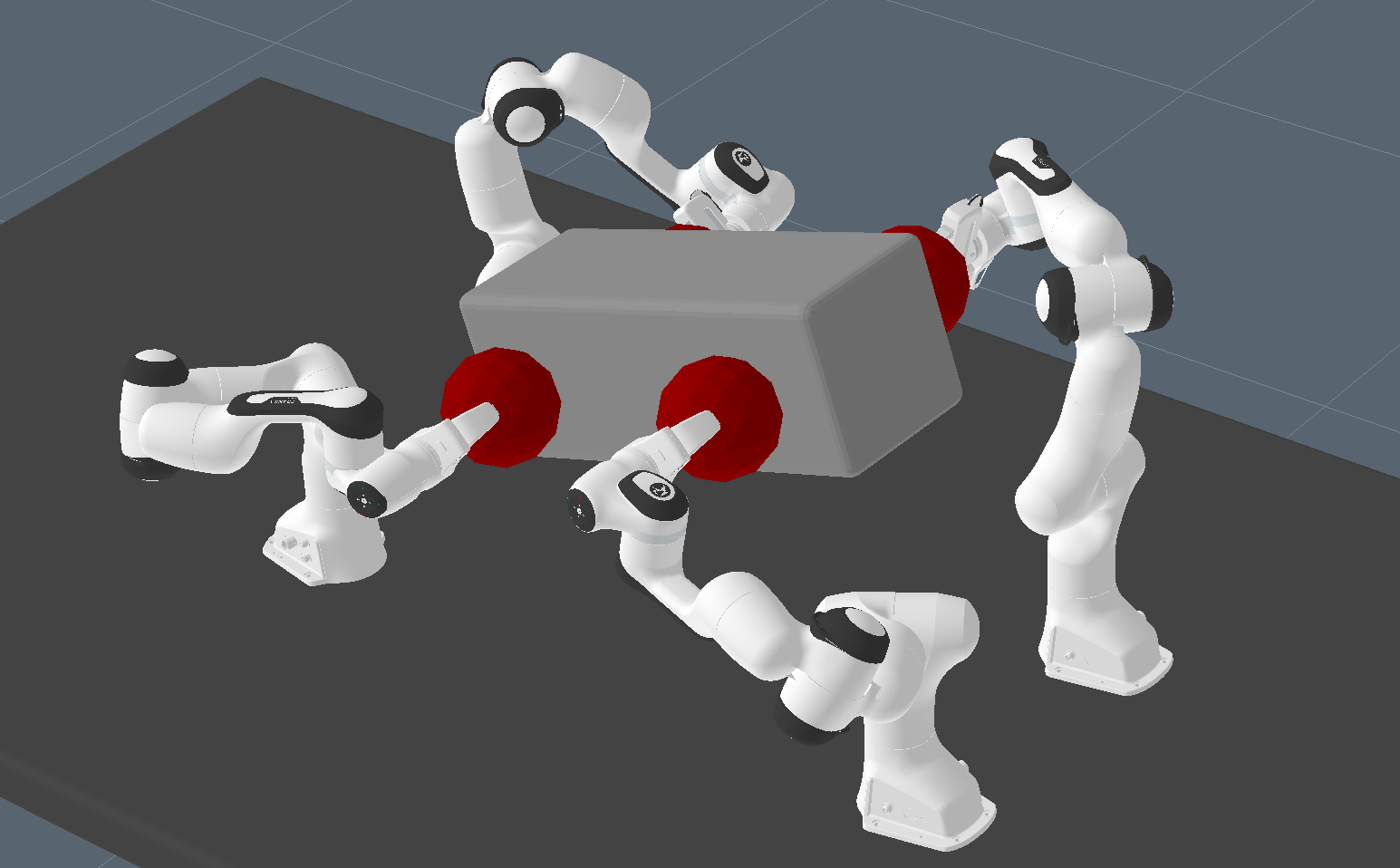}
        \caption{rotate horizontal object}
        \label{fig:subfig10000}
    \end{subfigure}
    \begin{subfigure}{0.238\textwidth}
        \centering
        \includegraphics[width=\linewidth]{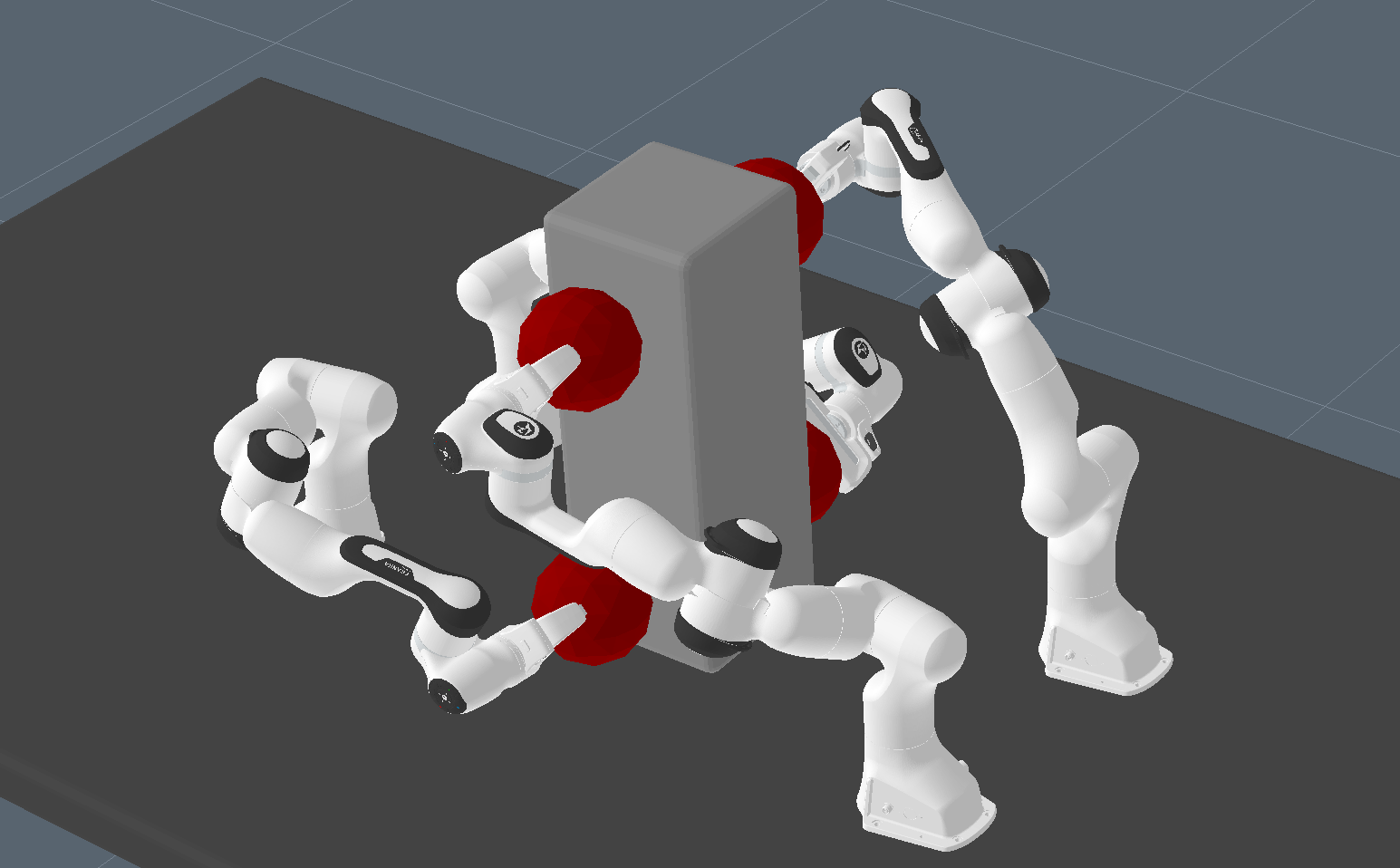}
        \caption{put it onto the ground}
        \label{fig:subfig20000}
    \end{subfigure}
    
    \vspace{0.2cm} % Add vertical space between rows

    \begin{subfigure}{0.238\textwidth}
        \centering
        \includegraphics[width=\linewidth]{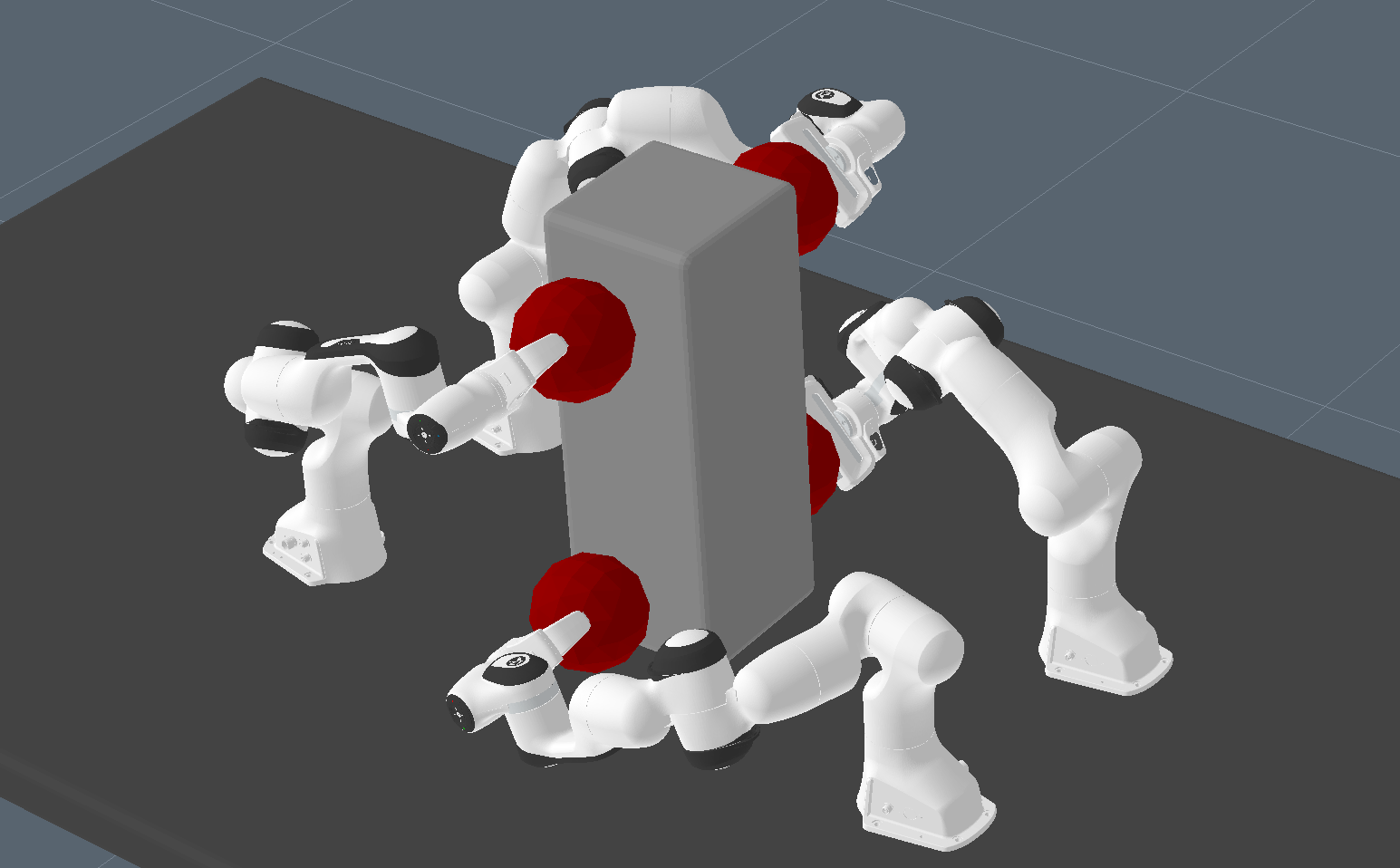}
        \caption{change contact points}
        \label{fig:subfig30000}
    \end{subfigure}
    \begin{subfigure}{0.238\textwidth}
        \centering
        \includegraphics[width=\linewidth]{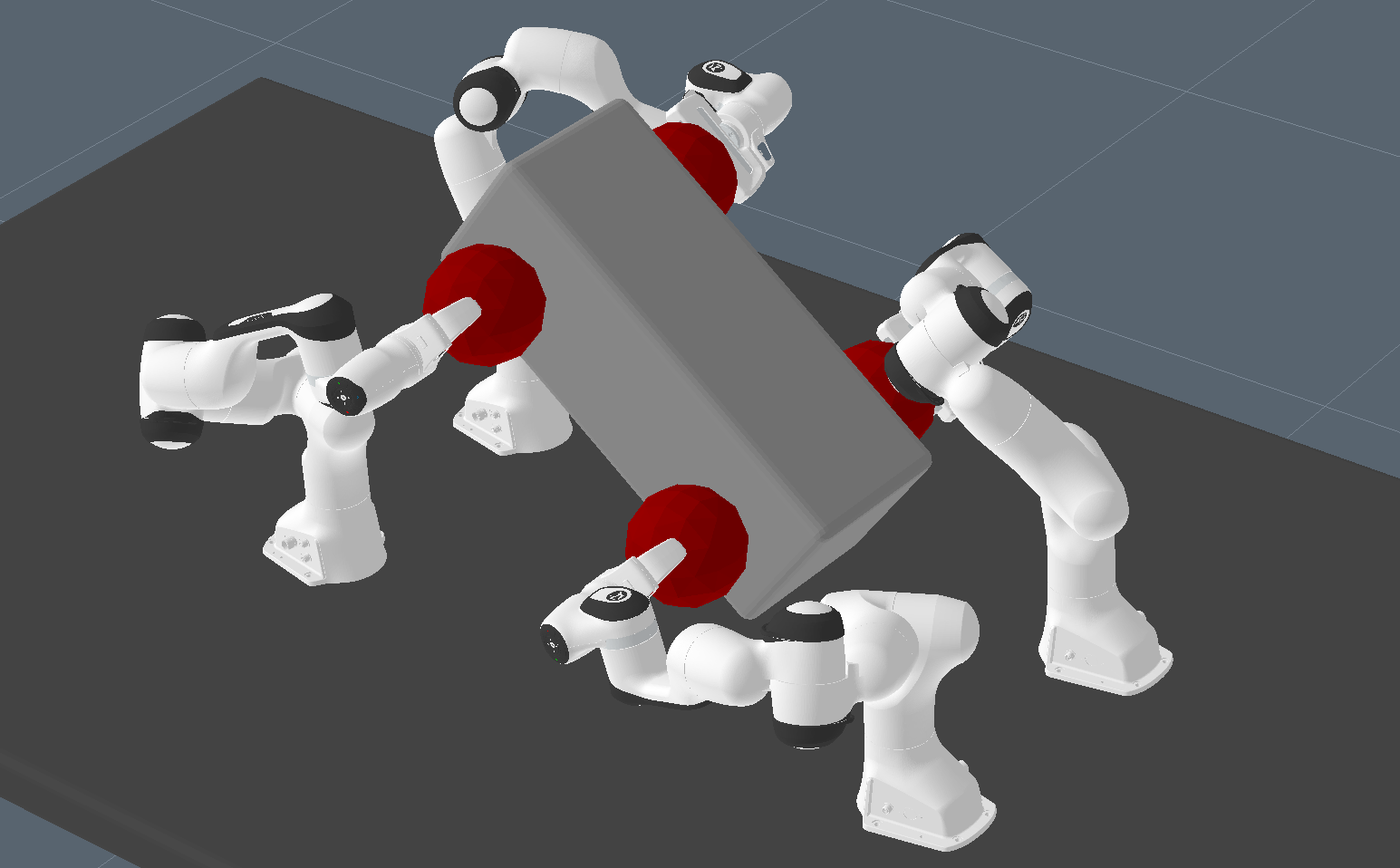}
        \caption{rotate for another $90 \deg$}
        \label{fig:subfig40000}
    \end{subfigure}

    \caption{Robot arms pick the object and reorient it in two stages. During this motion, all four robot arms must be synchronized.}
    \label{fig:exp4}
\end{figure}

Four robots that act like two pair of bimanual agents, where bimanual arms are $50 cm$ and agents are $1.2 m$ apart, reorient a box of size $30x30x80 cm$ in two stages: First, they make it upright by rotating to $90^\circ$, and after changing the contact points they rotate it by another $90^\circ$ (Fig. \ref{fig:exp4}). 
All four arms are required to be coordinated precisely to complete the task successfully, which our method achieves through optimization-augmented force control.

Our method completes all of these four tasks robustly, which have various settings and requirements. It is essential to integrate optimization-based motion planning with force control, because:

\begin{itemize}

\item \textbf{Without regulating the contact force, a reliable grasp cannot be achieved:} A firm grasp/contact is the de-facto requirement for the successful manipulation. Contact forces must be explicitly considered due to different object properties (e.g., mass, friction coefficient), and must be frequently regulated to keep the grasp stable. None of the tasks could be solved without the contact force regulation. 

\item \textbf{Without optimization-based motion planning, proper manipulation over a long-horizon cannot be achieved:} In our tasks, robots need to move over a long distance while keeping the transformation between end effectors constant; since robots have point contact with the object, any misalignment or misorientation leads to an undesired motion; and due to the underactuated and highly coupled nature of the system, it is nontrivial to fix these effects later. Thus, synchronization between arms is critical. In the absence of optimization, none of the tasks with multiple robots could be achieved reliably. Even for the first task, optimization allows for a more controlled motion. Apart from precise pose control, optimization is necessary for considering joint limits.

\item \textbf{Without force control, contact-rich manipulation tasks cannot be achieved:} A pure position control approach cannot adapt to the effects of contact forces, and it gets even more difficult when they act in multiple axes. If no force controller were available, single-arm box relocation and bimanual peg-in-hole tasks couldn't be completed.

\end{itemize}

\begin{figure}[h]
    \centering
    \includegraphics[width=0.45\textwidth]{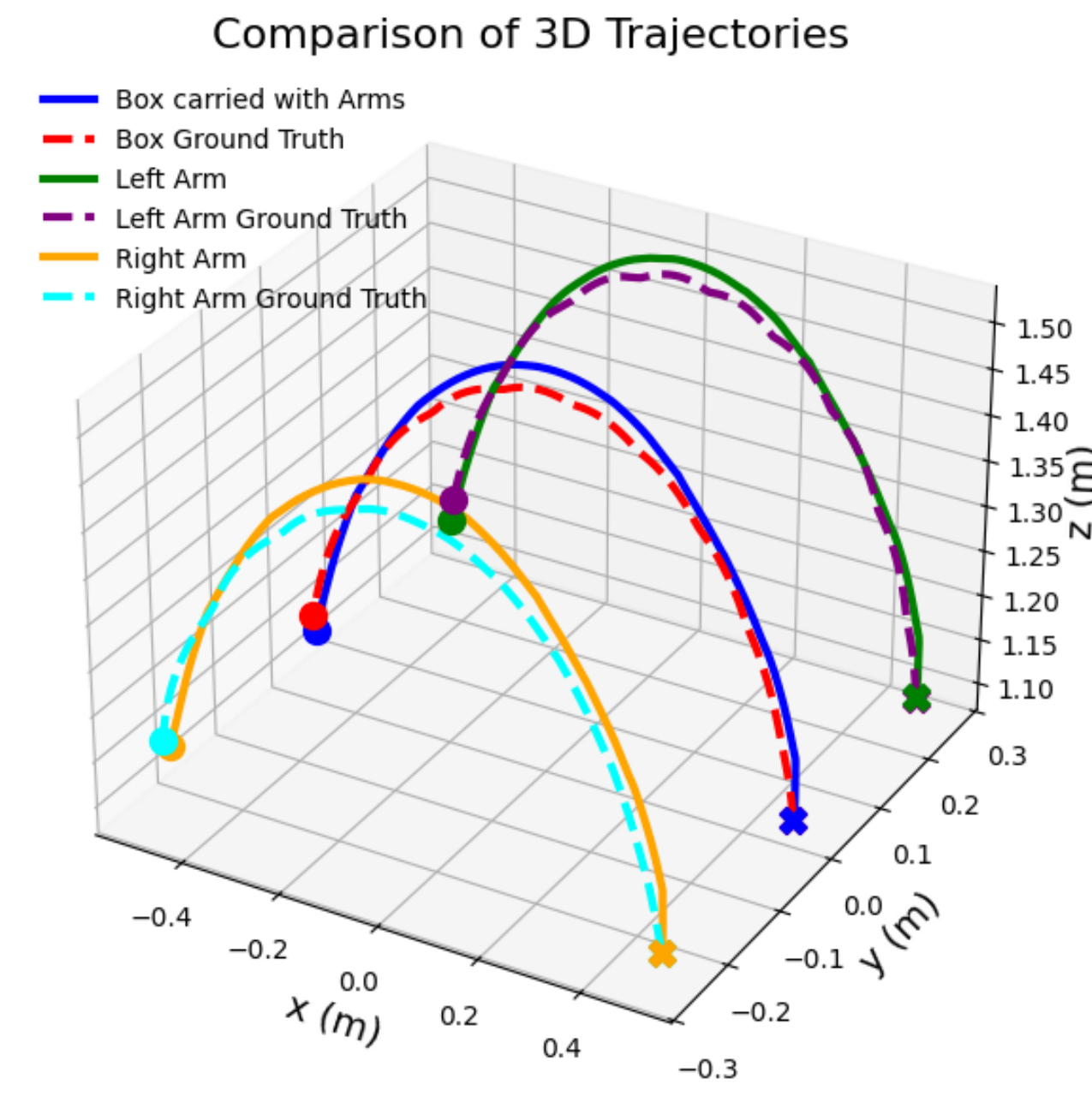}
    \caption{Trajectory of the system (Fig. \ref{fig:exp2}).}
    \label{fig:trajectory}
\end{figure}

\begin{figure}[h]
    \centering
    \includegraphics[width=0.45\textwidth]{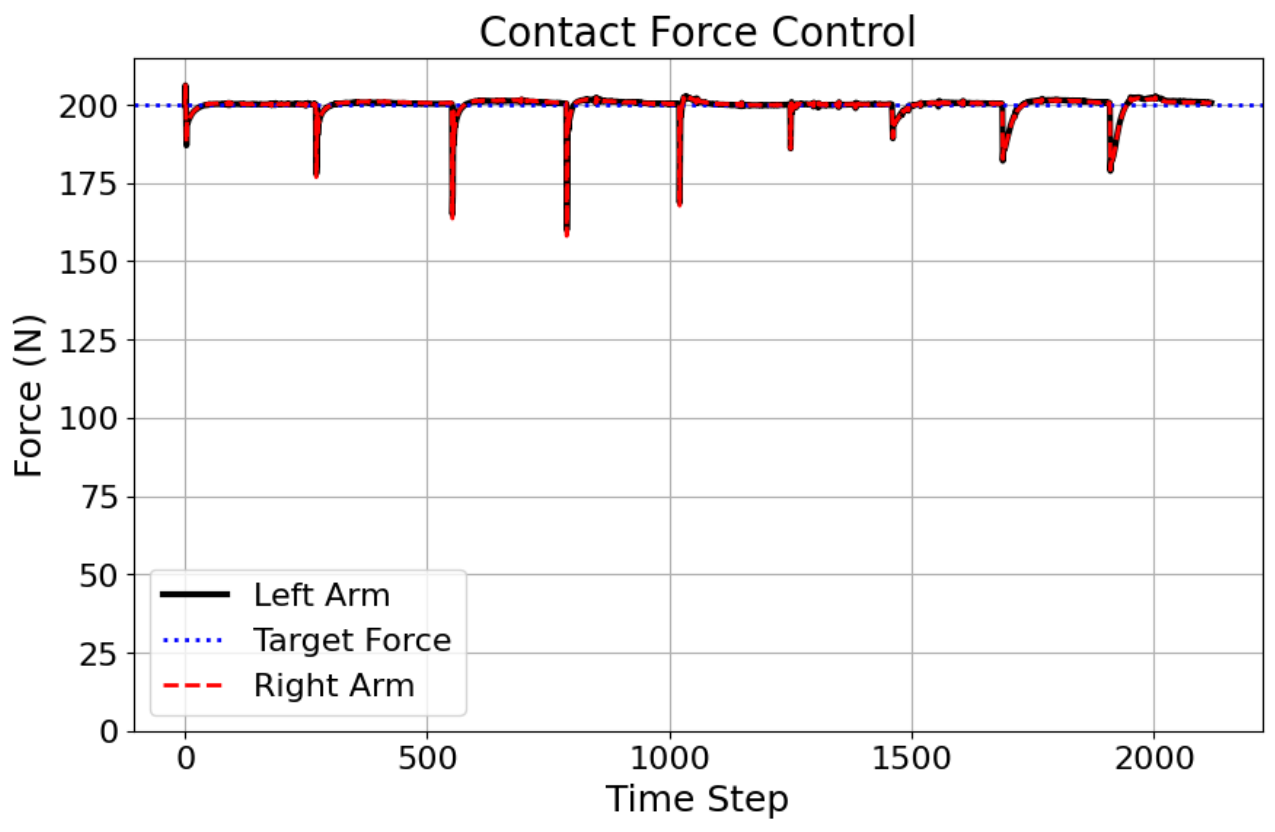}
    \caption{Contact forces over time (Fig. \ref{fig:exp2}).}
    \label{fig:contact_forces}
\end{figure}

\paragraph*{Practical Stability Analysis}
We assume that during the optimization-augmented force control the force direction and motion direction are orthogonal to each other, hence optimization controller must reach its target waypoints, and contact force controller must regulate the contact forces successfully. 
To test our intuition, we examined data from \textit{bimanual box transportation} case (Fig. \ref{fig:exp2}). 
Robot arms are commanded into the target positions with and without the box in between involded, using the optimization solution with 9 waypoints. Thus, the path is divided into 9 segments and robots are commanded to go to waypoints one after the other. Trajectories and contact forces are provided in Fig.~\ref{fig:trajectory} and Fig.~\ref{fig:contact_forces}, respectively. As we observe, at every switch for the next waypoint, contact forces experience a spike. 
Yet, the controller quickly stabilizes it, with root mean squared error (RMSE) of $2.80 N$ and $2.82 N$ for left and right arm, respectively. 
Similarly, the trajectory data is very close to 
the ground truth, for both robot arms and the box.
Since ground truth data is obtained by free space motion without the box involved, their durations are different. 
In order to compare trajectories, \textit{FastDTW} algorithm is utilized \cite{fastdtw}. After that, the RMSE values
of the left arm, right arm, and the box trajectories are obtained as $1.7 cm$, $1.6 cm$, and $2.1 cm$. Also, as we command the robot arms to go to one configuration after another, we consider 
the all 7 joint angle errors combined. 
RMSE values of the combined joint angles are $7.9^\circ$ for the left and $8.3^\circ$ for the right, which is around $1^\circ$ error per joint. 
In essence, our proposed controllers work in harmony, and targets can be reached robustly.

\section{DISCUSSION AND CONCLUSION}

In this work, we introduced a switching control framework that bridges the gap between optimization-augmented motion planning and force control, enabling robots to perform complex manipulation tasks with precision, adaptability, and robustness. Our framework leverages the strengths of both paradigms by decomposing tasks into subtasks tailored to specific control modes—pure optimization, pure force control, or a hybrid of both. This approach ensures that each subtask is executed using the most suitable controller, maximizing efficiency and task success rates. Through experimental validation, we demonstrated the versatility and effectiveness of our framework in addressing various manipulation challenges, and also verified its convergence behavior.

While our framework represents a significant step forward, there are opportunities for further 
improvement and exploration:
% and exploration

\begin{itemize}

\item Algorithmic Task Decomposition: Currently, task decomposition is performed manually, which limits scalability and generalizability. Automating this process using algorithmic methods, such as machine learning or rule-based systems, would enable the framework to dynamically identify and assign subtasks based on real-time observations and task requirements.

\item Experimental Validation: We showcased our method in a simulated environment with realistic physics, but future work will focus on extensive experimental validation of the proposed framework with real-world hardware.

\item Real-Time Replanning: While our framework supports iterative replanning, further research is needed to reduce computational overhead and enable real-time adaptation in highly dynamic environments. Also, additional sensory input (e.g., camera) is needed to decide when to replan, which we consider to investigate. 

\end{itemize}

This work lays the foundation for more advanced robotic systems capable of performing complex, contact-rich manipulation tasks in diverse and dynamic environments. By integrating optimization-augmented planning with force control and task decomposition strategy, our framework addresses a critical gap in robotic manipulation, enabling robots to operate more effectively in long-horizon manipulation scenarios with single or multiple robots.

\addtolength{\textheight}{-12cm}   % This command serves to balance the column lengths

%%%%%%%%%%%%%%%%%%%%%%%%%%%%%%%%%%%%%%%%%%%%%%%%%%%%%%%%%%%%%%%%%%%%%%%%%%%%%%%%

%%%%%%%%%%%%%%%%%%%%%%%%%%%%%%%%%%%%%%%%%%%%%%%%%%%%%%%%%%%%%%%%%%%%%%%%%%%%%%%%

%%%%%%%%%%%%%%%%%%%%%%%%%%%%%%%%%%%%%%%%%%%%%%%%%%%%%%%%%%%%%%%%%%%%%%%%%%%%%%%%

%%%%%%%%%%%%%%%%%%%%%%%%%%%%%%%%%%%%%%%%%%%%%%%%%%%%%%%%%%%%%%%%%%%%%%%%%%%%%%%%

\bibliographystyle{IEEEtran}  % IEEE-style bibliography
\bibliography{references}

\end{document}